# Detecting Harmful Short Videos in Toxic Diet Culture: A Governance-Ready Evidence Framework with Sparse Autoencoders

Wenli Zhang, Ph.D.
Iowa State University
3332 Gerdin, 2167 Union Drive
Ames, IA, USA 50011-2027
Email: wlzhang@iastate.edu

Shuang Geng, Ph.D.
Shenzhen University
1066 Xueyuan Road
Shenzhen, China 518055
Email: gs@szu.edu.cn

Jiaheng Xie, Ph.D.
University of Delaware
217 Purnell Hall
Newark, DE, USA 19716
Email: jxie@udel.edu

Zhonghan (Terry) Xiang
Iowa State University
3229 Gerdin, 2167 Union Drive
Ames, IA, USA 50011-2027
Email: xzh1120@iastate.edu

Sudha Ram
University of Arizona
McClelland Hall 430J, 1130 E. Helen St.
Tucson, Arizona 85721-0108
Email: sram@arizona.edu

Please send comments to Wenli Zhang at wlzhang@iastate.edu.

# Detecting Harmful Short Videos in Toxic Diet Culture: A Governance-Ready Evidence Framework with Sparse Autoencoders

## Abstract

Algorithmic content moderation is increasingly important for governing short-form video content, yet harmful-content detection research focuses primarily on prediction or limited explanations. This leaves a governance challenge underexplored: producing decision-linked evidence that supports the review and justification of moderation decisions. Using toxic diet culture as an empirical case, we conceptualize governance-ready evidence as computational traceability, evidence localization, policy grounding, and evidence inspectability. We develop the Sparse Autoencoder-based Policy-Aligned Contrastive Evidence (SPACE) framework in a pretrained SAE feature space derived from a large multimodal model. A global relevance gate and decomposable linear evidence head identify decision-linked evidence; segment supervision and within-video ranking localize this evidence; policy alignment maps it to platform guidelines; and Pareto-efficient selection yields a compact, task-specific evidence vocabulary. On 3,711 labeled TikTok videos, SPACE achieves the highest harmful-category Macro F1 among benchmarks across four families, at 0.8507. Its highest-scoring observation overlaps a ground-truth harmful span for 88.0% of harmful test videos, while its evidence representation retrieves the corresponding policy category for 87.0%. Stakeholder evaluations indicate that the SPACE evidence report is perceived as useful for understanding, reviewing, documenting, and responding to moderation decisions. The study contributes computational design science knowledge for embedding governance-ready evidence within predictive content-moderation systems.

# 1 INTRODUCTION

Short-form video platforms have become a major part of the digital information environment, through which millions of users encounter and engage with health-related content. These platforms can provide social connection and useful health information, but their algorithmically curated feeds can also repeatedly expose users to potentially harmful content, including messages that normalize extreme dieting, disordered eating, unhealthy weight-control practices, and appearance-based comparison (Paul, 2021). This challenge is particularly acute in toxic diet culture, where harmful messages are often embedded within seemingly ordinary food, fitness, beauty, and lifestyle videos and are therefore difficult to identify. Repeated exposure to such messages may shape how users perceive their bodies, evaluate eating and exercise behaviors, and respond to health information (Holland & Tiggemann, 2016). Prior research has linked social media exposure to body-image concerns and disordered-eating outcomes, while a recent study suggests that short-form video exposure can materially affect users' mental health (Xie et al., 2026). Recent litigation further illustrates the potential severity: two cases allege that algorithmic recommendations escalated from seemingly healthy content to extreme diet and body-image messages and contributed to anorexia among teenagers, resulting in inpatient treatment in one case and death by suicide in the other (Alter, 2025). Effective detection and governance are therefore important for limiting the large-scale amplification of toxic diet-related content and its potential effects on psychological well-being and health behavior.

For digital platforms, harmful-content detection is fundamentally a governance[1] issue. Harmful-content detection systems can lead to consequential moderation actions, including additional review, reduced visibility, restriction, required revisions, and removal, with important implications for viewers, creators, and platforms. Failure to detect harmful content may leave vulnerable users exposed, while decisions made without adequate supporting evidence can be difficult to review, justify, and communicate. Prior IS research highlights the accountability demands of algorithmic content moderation (Q. He et al., 2025) and shows that moderation actions can shape subsequent creator behavior (X. Zhang et al., 2026). Harmful-content detection systems must therefore provide more than accurate classifications. When platforms use model predictions to support moderation decisions, they also need *evidence* showing what supports the judgment and how that evidence relates to the applicable policy.

[1] Governance is the system of rules, processes, and structures used to direct, control, and hold an organized group or entity accountable.

Such evidence helps users understand the model's reasoning and supports informed moderation decisions. Interpretable machine learning typically focuses on whether people can understand the basis of a model's output (Rudin, 2019). In platform governance, however, evidence must support not only model understanding but also the moderation decision informed by the model. Research on moderation transparency similarly emphasizes decision-specific information about the moderated content, the applicable rule, and the roles of automated and human processes (Suzor et al., 2019). We define *governance-ready evidence* as decision-linked information that supports the broader governance function of content moderation. In automated short-video moderation, such evidence connects a moderation judgment to the computational basis of the prediction, the relevant content segment, and the applicable policy category, while making these relationships inspectable to the stakeholders who review or act on the decision. The central challenge is therefore not merely to explain a harmful-content prediction, but to *provide evidence that can support the governance decision it informs*.

Considering the information needed to support moderation decisions, we conceptualize governance-ready evidence for harmful-content detection as a multi-objective design problem comprising four interrelated requirements (Table 1). First, *computational traceability* requires the reported evidence to remain mathematically connected to the model prediction. Second, *evidence localization* identifies where decision-relevant harmful content occurs, particularly when harmful and benign material coexist within a video. Third, *policy grounding* connects this evidence to the policy under which the content is evaluated. Fourth, *evidence inspectability* ensures that the evidence is sufficiently focused and semantically meaningful for human review. These four requirements form an evidence chain that shows *where the relevant content appears, what computational evidence supports the judgment, which policy gives the evidence governance meaning, and whether the evidence can be feasibly inspected*. This chain translates platform governance requirements into design requirements for an algorithmic content moderation system. Simultaneously satisfying these four design requirements is non-trivial. These requirements are structurally interdependent: each provides context for the others and shapes what constitutes a satisfactory solution to them. *Computational traceability* requires the evidence presented for review to remain connected to the model's prediction, while *evidence inspectability* requires reducing a large internal feature space to evidence that a human can feasibly examine without losing that decision relevance. *Evidence localization* requires identifying the specific portion of a video that supports the prediction, while *policy grounding* requires that the same decision-relevant evidence acquire meaning under an external governance standard. A solution to any one requirement in isolation therefore does not establish governance-ready evidence. Reconciling these interdependent requirements within a coherent evidence chain presents a fundamental computational design challenge.

**Table 1. Governance Requirements and Corresponding Design Requirements for Automated Harmful-Video Moderation**

| Governance question | Design Requirement (DR) | Governance Standard and Literature | Implication for artifact design |
|---|---|---|---|
| Did the reported evidence actually contribute to the model prediction? | DR1. Computational traceability | (Padmanabhan et al., 2022; Rudin, 2019) | Explanatory evidence should remain mathematically connected to the prediction it supports. |
| Where is the evidence supporting the decision? | DR2. Evidence localization | (Chancellor et al., 2017; Li et al., 2026; Regulation (EU) 2022/2065, 2022, Article 17(3)(b); TikTok, 2024; Yang et al., 2026) | Evidence should be learned at the temporal level where harmful content occurs, and video-level judgments should be constructed from that localized evidence. |
| Under which external rule does the evidence become relevant for | DR3. Policy grounding | (Regulation (EU) 2022/2065, 2022, Article 17(3)(e); | Internal computational evidence should be connected to the external policy standard used to evaluate the content. |
| governance? | | Samory et al., 2026; Suzor et al., 2019) | |
| Can a human feasibly examine the evidence used by the system? | DR4. Evidence inspectability | (Q. He et al., 2025; Rudin, 2019; X. Zhang et al., 2026) | A large general-purpose feature space should be converted into a bounded, task-relevant evidence vocabulary whose semantic structure can be inspected. |

Existing research has developed important forms of evidence for harmful-video analysis, including interpretable semantic representations, policy-aware descriptions, and spatiotemporally grounded explanations (Li et al., 2026; Xie et al., 2026; Yang et al., 2026). However, these approaches largely address individual parts of the evidence problem and leave the four elements required for governance decisions disconnected: the observable content that raises concern, the internal computational basis of the prediction, the external policy used to evaluate the content, and the stakeholders responsible for reviewing moderation decisions. Crucially, this connection is not a simple aggregation of isolated outputs, but an interdependent evidence chain where each dimension contextualizes the others; without such cohesion, even an accurate prediction or plausible explanation leaves moderators uncertain about whether the presented evidence contributed to the model's judgment and how it relates to the governing policy. This gap raises a broader design question: *how can an automated harmful-content detection system produce evidence that is localized, computationally traceable, policy-grounded, and feasible for human inspection?*

Computational traceability presents a key challenge in addressing this question. Content moderation requires interpreting complex and context-dependent signals in videos, making the advanced representational capabilities of large multimodal models (LMMs) particularly valuable. However, the same complexity that enables these models to support difficult moderation tasks also makes their high-dimensional internal representations difficult to interpret. Addressing this challenge requires an automated detection system that can identify which features within an LMM's internal representations contribute to its prediction and make this evidence available for inspection. Sparse autoencoders (SAEs) offer a promising foundation for this purpose by decomposing complex model representations into sparse features, many of which exhibit recognizable semantic meaning (Cunningham et al., 2023; K. Zhang et al., 2025). SAEs may thus help transform an LMM's internal representations into interpretable evidence while preserving their connection to the underlying computation.

Semantically interpretable SAE features alone, however, are insufficient for platform governance. A feature may have a clear semantic meaning, yet still fail to identify the relevant video segment, show how it contributes to the model's prediction, or explain how that prediction relates to the applicable policy. Governance-ready evidence requires more than extracting interpretable representations. These representations must be organized into a focused evidence chain that clearly links the relevant video content, the model prediction made by the content moderation system, and the governing policy, allowing stakeholders to inspect how the decision was reached. This leads to a central design challenge: *how to transform interpretable internal representations of LMMs into a coherent evidence chain that can support governance decisions*.

We address this challenge by developing and evaluating the Sparse Autoencoder-based Policy-Aligned Contrastive Evidence (SPACE) framework. Using toxic diet culture as the empirical setting, we examine whether SPACE can jointly satisfy the four governance-derived requirements while maintaining strong predictive performance. To address the computational challenge created by these interdependent design requirements, SPACE learns task-specific feature relevance over a pretrained SAE feature space through a global gate, producing a shared gated representation through which the four requirements can shape a coherent evidence chain. We evaluate SPACE on a real-world dataset to assess both its predictive performance and its ability to satisfy these four governance-derived requirements. Additional evaluations with content creators and platform moderation professionals indicate that the governance-ready evidence generated by SPACE is perceived as useful for understanding, reviewing, documenting, and responding to moderation decisions.

Following the design science paradigm (Gregor & Hevner, 2013; Hevner et al., 2004), this study contributes design knowledge on how platform-governance requirements can shape automated harmful-content detection. We position the study as computational design science, which emphasizes domain-driven computational innovation with business and societal impact (Rai, 2017), and as Type I machine-learning research, in which the characteristics of an important business or societal problem guide the development and evaluation of a machine learning method (Padmanabhan et al., 2022). Our central contribution is to conceptualize and instantiate *governance-ready evidence* as a design objective comprising localization, computational traceability, policy grounding, and inspectability. This perspective extends harmful-content research beyond whether a system can generate accurate predictions and meaningful explanations to whether it can establish the evidentiary relationships needed when model predictions inform organizational decisions. By implementing these requirements in SPACE, the study demonstrates how a governance context can shape both the architecture and evaluation of a machine learning framework.

In practice, this study clarifies how automated detection systems can better support stakeholders who make or receive moderation decisions. Governance-ready evidence helps moderators identify the relevant content, inspect the computational basis of a prediction, connect that evidence to platform policy, and document the decision. It also gives content creators more specific information about what triggered a moderation action and how the concern relates to the applicable policy. More broadly, toxic diet culture represents one instance of a larger class of content governed through platform community guidelines. Platforms also regulate content involving self-harm, threats and violence, and other harmful behaviors. Although the substantive signals differ across these domains, automated moderation raises the same governance questions: where does the decision-relevant evidence occur, what computational evidence supports the judgment, which community guideline provides the policy basis, and can that evidence be feasibly inspected by human reviewers? Accordingly, the four requirements of evidence localization, computational traceability, policy grounding, and evidence inspectability provide a governance framework for automated moderation across diverse forms of policy-regulated content. This framework extends the study beyond toxic diet culture by specifying the evidence capabilities that automated moderation systems should provide whenever platform decisions are governed by community guidelines.

# 2 RELATED WORK

## 2.1 Governance Requirements for Decision-Linked Moderation Evidence

Research on content moderation transparency identifies core informational elements involved in moderation decisions, including the content being evaluated, the applicable platform rule, and the respective roles of automated and human processes (Suzor et al., 2019). Research on algorithmic moderation further shows that model outputs become part of the organizational processes through which platforms make and implement governance decisions (Q. He et al., 2025; X. Zhang et al., 2026). Building on this literature, we conceptualize an automated moderation decision as linking four elements: the content being evaluated, the model prediction produced by the detection system, the platform policy under which the content is assessed, and the stakeholder who must examine or act on the resulting information. The four design requirements introduced in the Introduction (Table 1) can therefore be derived from the informational relationships among these elements.

Figure 1 links each of the four design requirements to a distinct evidentiary relationship within a moderation decision. (*i*) The first relationship connects the model prediction to the computation that produced it. It concerns whether reviewer-visible evidence is intrinsic to the predictive model and directly connected to its output (Rudin, 2019). We refer to this requirement as *computational traceability*. (*ii*) The second relationship connects the model prediction to the content being evaluated. In short-form video, harmful and benign material can coexist within the same video, so a useful evidentiary record should identify where the content relevant to the judgment occurs (Li et al., 2026). We refer to this requirement as *evidence localization*. (*iii*) The third relationship connects the evidence to the platform policy used to evaluate the content. Platform moderation is organized around community guidelines and related policy categories. Evidence therefore needs a policy basis that explains why the identified content is relevant to moderation. This requirement is consistent with transparency research that emphasizes communicating the rule implicated by a moderation decision (Suzor et al., 2019). We refer to this requirement as *policy grounding*. (*iv*) The fourth relationship connects the evidence to the stakeholder who must examine it. Evidence that is too diffuse, semantically opaque, or extensive provides limited support for review even when it contains decision-relevant information (Rudin, 2019). We refer to the requirement that evidence be organized in a form that permits feasible human examination as *evidence inspectability*.

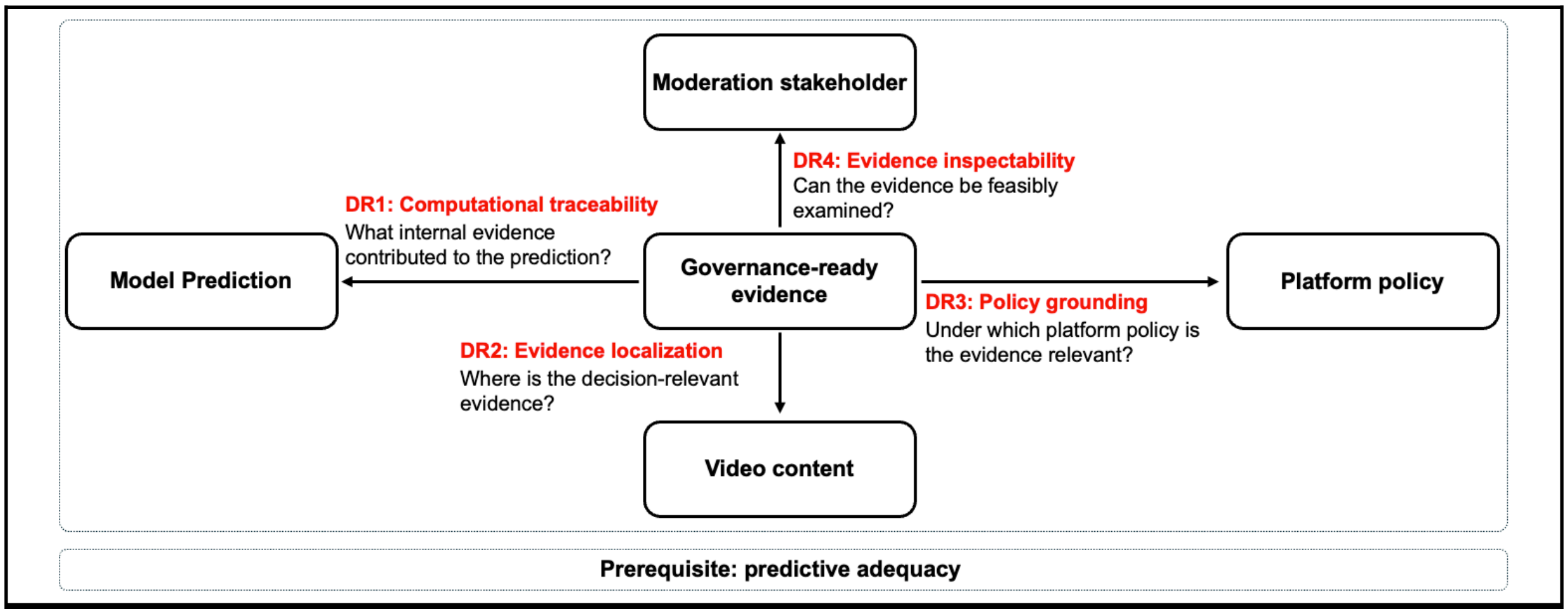

**Figure 1. Governance-Ready Evidence and Design Requirements for Short-Video Moderation**

Note: DR = Design requirement. Arrows represent evidentiary relationships required for moderation review, not a sequential moderation workflow. Predictive adequacy is treated as a prerequisite for the evidentiary function of an algorithmic content moderation system.

These four relationships define the evidentiary function examined in this study: localizing the basis of a moderation judgment in the content, tracing the prediction to the underlying model computation, making the evidence inspectable to stakeholders, and grounding it in platform policy. They provide a focused basis for assessing whether an algorithmic content moderation system can support moderation review beyond category prediction. Predictive adequacy remains a necessary condition because evidentiary support is of limited value when the underlying predictions are unreliable. Broader moderation concerns, including fairness, appeals, recourse, enforcement design, workload, and organizational accountability, operate at distributional, procedural, or organizational levels. This study focuses on governance-ready evidence as the information available to examine and interpret an automated judgment for a moderation decision.

The four governance requirements guide the remainder of the review. Section 2.2 examines how harmful video research has evolved from category prediction toward richer forms of evidence. Section 2.3 compares interpretable and policy-aware moderation approaches based on the evidence they provide and the relationships they reveal among content, computation, policy, and human review, highlighting both existing capabilities and unresolved gaps. Section 2.4 then examines how computational evidence can remain linked to model predictions while still being accessible for human inspection.

### 2.2 Evidence Capabilities in Harmful-Video Detection

Research on harmful-content detection has traditionally centered on prediction: given a piece of social-media content, the system determines whether it belongs to a harmful category. This capability is a necessary foundation for automated moderation because problematic content must first be identified with sufficient accuracy. Prior work has therefore developed increasingly capable models for detecting harmful signals across text, images, and video. In the eating-disorder context, Chancellor et al. (2017) developed multimodal classifiers to identify pro-eating-disorder content on social media. More recent IS research has extended automated detection to settings such as hate speech and has shown that modeling choices, training labels, and data design can materially affect detection outcomes (Kim et al., 2025; Wei et al., 2025). These studies establish predictive adequacy as an essential baseline for any automated moderation system.

For short-form videos, a content-level label or score provides limited information about the evidence supporting a moderation judgment. Harmful signals can be multimodal, context-dependent, and temporally concentrated, so a correct video-level prediction may still leave the decision-relevant content unidentified (Chancellor et al., 2017; Li et al., 2026; Yang et al., 2026). This limitation has shifted harmful-video research toward richer forms of evidence about what in the video supports the judgment.

A related stream enriches video analysis with semantically meaningful representations. Xie et al. (2026) provide an example in the short-video and mental-health context. Their knowledge-guided neural topic model incorporates clinically grounded knowledge into the representation of short-form videos and identifies interpretable topics that connect video content with viewers' mental-health outcomes. Its contribution extends beyond improved prediction by showing that a computational model can expose meaningful semantic structure in video content and that domain knowledge can shape the concepts used to interpret that content. This work establishes semantic evidence as a legitimate output of video analysis beyond a category label alone.

Prediction and semantic representation establish two important capabilities for harmful-video analysis: identifying problematic content and characterizing its substantive meaning. However, existing work leaves open how the evidence supporting an individual moderation judgment can be localized, traced to the predictive computation, and interpreted under the applicable policy. Section 2.3 examines how existing evidence-generation approaches address this gap.

### 2.3 Interpretability and Explanation in Video Moderation

Existing literature on interpretable video analysis and content moderation provides several forms of evidence beyond a category prediction. These approaches differ, however, in what they expose to a stakeholder and how that evidence relates to the content moderation decision. Table 2 compares prior approaches by the evidence they expose and by the four governance requirements derived in Section 2.1. This capability-based view clarifies which relationships each approach makes visible in the moderation process. Prior work may provide individual elements of governance-ready evidence; the unresolved issue is whether these elements can be jointly supported within the same predictive artifact: governance-ready evidence is a multi-objective design problem; strength on one requirement does not substitute for the others.

**Table 2. Evidence Capabilities of Interpretable and Policy-Aware Moderation Approaches**

| Evidence family and representative work | Evidence exposed / provenance | Governance Requirements | | | |
|---|---|---|---|---|---|
| | | Evidence Localization | Computational Traceability [a] | Policy Grounding [b] | Evidence Inspectability |
| Semantic interpretation: Xie et al. (2026) | Clinically informed semantic topics learned within the video representation | Primarily video-level | Semantic topics are integrated into the learned representation, but are not exposed as feature contributions composing the prediction | Clinical/domain knowledge informs the representation; platform moderation policy is not explicit | Compact, semantically meaningful topics |
| Post-hoc attribution: Ribeiro et al. (2016); Lundberg and Lee (2017) | Input-feature contributions estimated from model behavior | Depends on whether the input representation preserves temporal or spatial structure | Attribution estimates the contribution of input features to the output, while the model's internal latent computation remains outside the exposed evidence | Platform policy is not explicit | Inspectability depends on the input representation and number of attributed features |
| Generated explanations: Lang et al. (2026) | Human-readable explanations learned through explanatory supervision | Self-grounded content cues | Explanation is generated to account for the decision, but does not expose internal feature contributions composing the prediction | Platform policy is not explicit | Human-readable natural-language explanation |
| Grounded explanations: K. Li et al. (2026) | Evidential clues verified against observable video content | Spatiotemporal grounding | Verified clues support the detection and explanation process, but the primary evidence is not a decomposition of the final score into internal latent contributions | Platform policy is not explicit | Grounded and human-readable evidence |
| Structured multimodal evidence: Yadav and Singh (2026) | Visual, audio, textual, and contextual evidence retained for the decision process | Content- and modality-level evidence | Explicit evidence participates in the decision process, while internal latent contributions remain largely unexposed | Platform policy is not explicit | Structured multimodal evidence record |
| Rule-sensitive moderation: Samory et al. (2026) | Applicable community rule used in moderation | Primarily content-level | Prediction is conditioned on explicit community rules; internal feature contributions are not the primary evidence exposed | Explicit community-rule grounding | Explicit and inspectable rule identification |
| Policy-aware video interpretation: Yang et al. (2026) | Policy-aware video caption used as an intermediate representation | Fine-grained content description | Policy-aware caption participates in the moderation pipeline, but latent feature contributions to the final judgment are not the primary evidence exposed | Explicit platform-policy grounding | Human-readable policy-aware caption |

Note: [a] Computational traceability concerns whether reviewer-visible evidence can be connected to the computation producing the prediction. [b] Policy grounding concerns an explicit external moderation rule or platform policy; clinical or domain knowledge is treated as a distinct source of semantic structure. The comparison is limited to evidentiary capabilities relevant to Section 2.1.

Semantic approaches expose meaningful content concepts, while post-hoc attribution methods identify influential inputs (Lundberg & Lee, 2017; Ribeiro et al., 2016; Xie et al., 2026). These approaches improve semantic or attributional visibility, but they do not jointly connect localized video evidence, internal feature contributions that compose the prediction, and the external policy governing the decision.

Generated and grounded explanation approaches produce evidence that is easier to inspect: LEAF generates natural-language explanations, Yadav and Singh preserve structured multimodal evidence, and MATCH verifies explanatory clues against spatiotemporal video evidence (Lang et al., 2026; Li et al., 2026; Yadav & Singh, 2026). These approaches strengthen inspectability and, particularly for MATCH, evidence localization; however, their generated or verified explanations do not expose an exact decomposition of the final prediction into internal feature contributions, and platform policy remains outside the same evidence structure.

Policy-aware approaches address another part of the problem. ModQ conditions moderation on explicit community rules, UNIVID generates policy-aware video representations, and IPS uses policy-specific process supervision for short-video moderation (Liu et al., 2026; Samory et al., 2026; Yang et al., 2026). These approaches show that external rules and policy-specific reasoning can be incorporated into moderation, but they do not jointly connect the governing policy to localized internal evidence that composes the final prediction.

Inspectability becomes especially consequential when evidence is drawn from internal model representations. Such evidence must remain computationally connected to the prediction while being sufficiently bounded and semantically coherent for human review. This creates an internal-representation problem that Section 2.4 examines.

Overall, prior work has established several governance capabilities across different evidentiary objects and representational layers, but these capabilities are typically pursued in isolation. This fragmentation leaves a multi-objective design challenge: evidence localization, computational traceability, policy grounding, and evidence inspectability are structurally interdependent because each shapes the context in which the others become meaningful. Localized evidence must remain connected to the judgment; computational evidence must remain inspectable; policy grounding must apply to the evidence supporting the judgment; and feature reduction must preserve decision relevance. Coordinating these dependencies while maintaining predictive adequacy constitutes a coupled computational design problem.

### 2.4 Sparse Internal Representations and the Downstream-Utility Problem

Sparse autoencoders (SAEs) provide one candidate foundation for the internal-representation problem that follows from computational traceability and inspectability. Hidden states in LLMs and LMMs are dense and high-dimensional, making individual dimensions difficult to interpret directly. SAEs learn sparse latent dictionaries that can expose semantically interpretable directions in model activations (Cunningham et al., 2023). This capability has been extended to LMMs, where automatic interpretation identifies recognizable semantic structure in multimodal representations (K. Zhang et al., 2025). These features provide candidate internal evidence for harmful-video moderation. Recent work also identifies important limits: SAE features can vary across training runs, and video SAEs exhibit trade-offs among reconstruction, temporal coherence, discrimination, and interpretability (Dokme & Vishwanath, 2026; Song et al., 2026). These findings separate the quality of a sparse representation from its usefulness for a specific downstream decision task.

For harmful-video moderation, the downstream question is which SAE features should become decision-relevant evidence. A pretrained SAE is optimized for reconstruction under sparsity, so its feature dictionary is not selected for the moderation task. Its objective does not identify which features distinguish harmful from benign videos, locate the relevant portion of a video, or support interpretation under a platform policy. A large SAE dictionary may therefore contain many meaningful features whose relevance to moderation is unknown. The downstream problem is to determine which internal features should participate in the task-specific evidence produced by the detection system.

Evidence on downstream SAE utility remains mixed. Gallifant et al. (2025) report strong classification and transfer performance using SAE-derived features, whereas Kantamneni et al. (2025) find that SAE-based probes do not consistently outperform strong non-SAE baselines across a broad set of probing tasks. These results make downstream adaptation and task-specific evaluation central design issues.

Gated SAEs improve sparse representation learning by separating feature activation from magnitude estimation during SAE training (Rajamanoharan et al., 2024). The downstream selection problem examined here begins after SAE training and concerns which fixed SAE features should support the focal governance task.

This issue is particularly important for governance-ready evidence. A content moderation system must identify the internal features contributing to a harmful-content prediction, localize the supporting content within the video, connect that evidence to the applicable policy category, and present it in a form feasible for stakeholder inspection, all while maintaining predictive adequacy. Existing harmful-video research provides complementary capabilities in prediction, semantic interpretation, grounded explanation, multimodal evidence, and policy-aware moderation. SAE research provides a plausible internal-representation foundation for this multi-objective challenge. Its governance value depends on task-specific adaptation that coordinates localization, computational traceability, policy grounding, and inspectability with predictive adequacy. This motivates our research question (RQ): *How can automated harmful short-video detection be designed to produce governance-ready evidence for platform moderation decisions while maintaining predictive adequacy?* This research question positions governance-ready evidence as the design objective guiding the artifact design and evaluation that follow.

## 3 RESEARCH DESIGN

To address the research question, we develop SPACE as an integrated predictive framework designed to instantiate a cohesive evidence chain across all four requirements. Their structural interdependence creates a computational challenge: each requirement must be satisfied while preserving the evidentiary relationships on which the others depend. SPACE addresses this challenge by learning task-specific relevance over the pretrained SAE feature space to form a shared gated representation. Through joint multi-objective training, SPACE uses contrastive localization, policy grounding, and sparsity objectives to shape the global relevance gate over the pretrained SAE feature space. The resulting shared gated representation is then processed by a linear evidence head, which preserves computational traceability.

The computational design follows the dependency among the four design requirements. SPACE first constructs decomposable evidence that remains mathematically connected to the prediction, then constrains where that evidence should appear, relates the evidence to the applicable platform policy, and converts the underlying feature space into a bounded evidence vocabulary that can be inspected. These mechanisms are learned jointly with the prediction task, operationalizing governance-ready evidence as a multi-objective design problem.

Let $V_i = \{x_{i,1}, ..., x_{i,T}\}$ denote video $i$, represented by $T$ temporally ordered sampled observations, each treated as one model segment, and let $y_i \in Y$ denote its content category. For harmful videos in the training data, temporal annotations identify continuous harmful time spans; a sampled segment receives a harmful label when the timestamp of its observation falls within an annotated harmful time span. These temporal annotations are used only as training supervision; at inference, SPACE localizes evidence from its learned segment-level evidence scores without requiring temporal annotations. SPACE begins from the internal representation of a fixed pretrained LMM $F$. A fixed pretrained SAE $E$ transforms the representation of each segment into a sparse latent feature vector:

$$z_{i,t} = E\left(F(x_{i,t})\right) \in R^{K} \quad (1)$$

The pretrained LMM and SAE remain unchanged, and SPACE learns task-specific evidence mechanisms over the pretrained SAE feature space. SPACE learns a global feature-relevance vector

$$g = \sigma(\alpha) \in (0, 1)^{K} \quad (2)$$

and forms the gated representation

$$\tilde{z}_{it} = g \odot z_{it} \quad (3)$$

where $\odot$ denotes elementwise multiplication. The resulting $\tilde{z}_{i,t}$ is the shared gated representation used across the four evidence mechanisms. Because the same gate is shared across videos, it represents task-level feature relevance across the dataset. Section 3.1 uses this representation to construct decomposable evidence and establish computational traceability. Section 3.2 constrains the temporal location of that evidence. Section 3.3 grounds the evidence in platform policy. Section 3.4 learns and selects an inspectable evidence vocabulary. Section 3.5 combines these mechanisms in a joint learning objective. Figure 2 summarizes the research design.

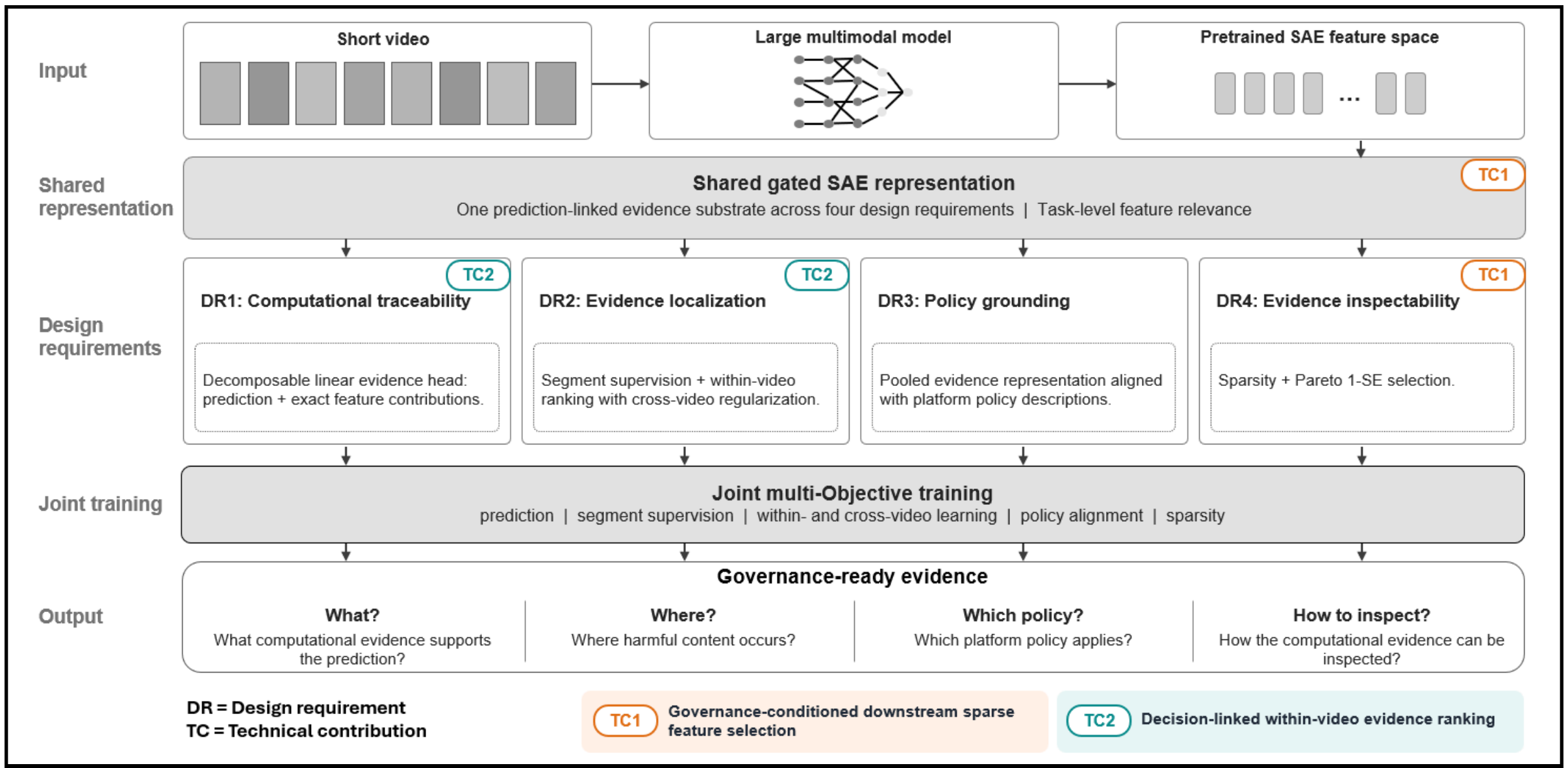

**Figure 2. SPACE Research Design**

### 3.1 DR1. Computational Traceability: Decomposable Evidence Construction

Computational traceability requires the evidence exposed for review to remain mathematically connected to the prediction it supports. SPACE implements this requirement through a linear evidence head applied directly to the shared gated representation $\tilde{z}_{i,t}$. For each segment $t$ of video $i$, the model computes category-specific evidence scores as

$$e_{it} = W\tilde{z}_{it} + b \tag{4}$$

where $W \in R^{C\times K}$ and $b \in R^{C}$ are learned parameters, and $e_{i,t,c}$ denotes the evidence that segment $t$ provides for category $c$.

Decision-relevant evidence may be concentrated in only a small portion of a video. SPACE therefore aggregates segment-level evidence using the strongest segment for each category:

$$e_{i,c} = \max_{t\in\{1,\dots,T\}} e_{i,t,c} \tag{5}$$

The resulting category scores are converted to video-level class probabilities through a softmax function, and the category with the largest probability determines the prediction. This aggregation also identifies the segment supplying the strongest evidence for each category. Section 3.2 subsequently uses temporal supervision to concentrate these evidence scores in harmful segments during training.

The linear evidence head provides an exact decomposition of each segment-category score. The contribution of SAE feature $k$ to category $c$ in segment $t$ is

$$\phi_{i,t,c,k} = W_{c,k}\tilde{z}_{i,t,k}, \qquad e_{i,t,c} = b_c + \sum_{k=1}^{K} \phi_{i,t,c,k} \tag{6}$$

Thus, the feature contributions reported as evidence are the same quantities that compose the model's category score. SPACE can trace a video prediction to the segment supplying its strongest category evidence and, within that segment, to the individual sparse features contributing to the score. Computational traceability in SPACE therefore extends from the video prediction to the strongest segment-category score and the gated SAE feature contributions that compose that score.

### 3.2 DR2. Evidence Localization: Within-Video Evidence Learning

Evidence localization requires the decision-linked evidence defined in Section 3.1 to concentrate on the portions of a video that contain the harmful content. Video-level classification alone cannot impose this requirement because a correct prediction may still be driven by a benign segment or by video-level characteristics correlated with the harmful category. SPACE therefore uses training-set temporal annotations to supervise where category evidence should appear.

The first localization signal is segment-level classification. During training, each sampled segment whose observation timestamp falls within an annotated harmful time span inherits the video's harmful category; all remaining sampled segments are assigned the Non-Toxic category 0. The segment-level classification loss is applied to the same category evidence scores $e_{i,t,c}$ used to construct the video-level prediction in Section 3.1. This provides direct supervision over which category each temporal unit should support.

Segment-level classification treats temporal units independently. SPACE additionally exploits the fact that harmful and benign segments can coexist within the same video. For each segment, we define its harmfulness score as the strongest harmful-category evidence relative to its benign evidence:

$$s_{i,t} = \max_{c>0} e_{i,t,c} - e_{i,t,0} \tag{7}$$

where category 0 denotes benign content. A larger $s_{i,t}$ indicates stronger evidence for harm relative to benign content.

For each harmful video $i$, let $H_i$ denote sampled segments whose observation timestamps fall within annotated harmful time spans and $B_i$ the remaining sampled segments. SPACE requires harmful segments to carry stronger decision evidence than benign segments from the same video:

$$L_{local} = Avg_{i,\, t^{+} \in H_i,\, t^{-} \in B_i} log\left[1 + exp\left(s_{i,t^{-}} - s_{i,t^{+}}\right)\right] \quad (8)$$

The loss decreases as $s_{i,t^{+}}$ exceeds $s_{i,t^{-}}$. It therefore imposes a relative localization constraint on the evidence scores that participate directly in prediction. The within-video objective builds on pairwise ranking formulations used in weakly supervised video analysis (Sultani et al., 2018) and applies them to a different computational object: the decomposable decision-evidence scores defined in Section 3.1. Comparing segments within the same video is particularly useful because they share creator identity, visual style, topic, and production context while differing in whether they contain the annotated harmful content. The comparison consequently directs learning toward the local evidence associated with harm and reduces reliance on video-level nuisance characteristics.

SPACE also includes a cross-video contrastive objective as an auxiliary regularizer over the gated SAE representations. Let $y_{i,t}$ denote the segment-level category defined above. For an anchor segment $(i, t)$, we sample a positive segment $(j, u)$ from a different video such that $y_{j,u} = y_{i,t}$, and define $N(i, t)$ as the set of sampled segments with $y_{k,v} \neq y_{i,t}$. The cross-video InfoNCE loss is

$$L_{cross} = - \frac{1}{|A|} \sum_{(i,t)\in A} log \frac{exp\left(sim\left(\tilde{z}_{i,t}, \tilde{z}_{j,u}\right)/\tau\right)}{exp\left(sim\left(\tilde{z}_{i,t}, \tilde{z}_{j,u}\right)/\tau\right) + \sum_{(k,v)\in N(i,t)} exp\left(sim\left(\tilde{z}_{i,t}, \tilde{z}_{k,v}\right)/\tau\right)} \quad (9)$$

where $A$ is the set of anchor segments in the training batch, $\tau$ is the temperature parameter, and $sim(\cdot, \cdot)$ denotes cosine similarity. The objective encourages segments with the same category from different videos to have similar gated representations while separating segments with different categories. Its role is auxiliary: $L_{cross}$ regularizes category-level structure across videos, whereas the within-video objective $L_{local}$ determines where decision evidence should be concentrated.

Segment supervision anchors category evidence to annotated harmful segments, while within-video ranking sharpens the separation between harmful and benign segments within the same video. The resulting evidence remains decision-linked through the scores defined in Section 3.1 and becomes temporally grounded. Section 3.3 aligns the shared gated representation with category-level policy semantics.

### 3.3 DR3. Policy Grounding: Evidence-Policy Alignment

Policy grounding requires the computational evidence used by SPACE to be interpretable against the external policy standard governing the moderation decision. For each category $c \in Y$, we construct a policy anchor from the detailed natural-language description of the corresponding platform guideline. Let $q_c \in R^{d_p}$ denote its fixed, L2-normalized text embedding produced by a pretrained text encoder.

SPACE connects the shared gated representation to these policy anchors through a learned projection. We first summarize the gated representations across the video:

$$\bar{z}_i = \frac{1}{T}\sum_{t=1}^{T} \tilde{z}_{i,t} \tag{10}$$

and project the resulting representation into the policy embedding space:

$$v_i = \frac{P_\theta(\bar{z}_i)}{\|P_\theta(\bar{z}_i)\|_2} \tag{11}$$

where $P_\theta(\cdot)$ is a learned policy projector and $v_i \in R^{d_p}$ is the normalized policy-aligned evidence representation.

The policy-alignment objective encourages $v_i$ to be closest to the policy anchor associated with the ground-truth category:

$$L_{policy} = -\frac{1}{N}\sum_{i=1}^{N} log\frac{exp\left(sim(v_i, q_{y_i})/\tau_p\right)}{\sum_{c\in Y} exp\left(sim(v_i, q_c)/\tau_p\right)} \tag{12}$$

where $sim(\cdot,\cdot)$ denotes cosine similarity and $\tau_p$ is the policy-alignment temperature.

This objective complements the classification mechanism in Section 3.1. The linear evidence head learns whether the gated features support a particular content category, while $L_{policy}$ constrains the representation formed from those features to remain semantically consistent with an external description of the corresponding platform policy. Policy grounding is therefore imposed on a pooled representation derived from the same shared gated representation used for prediction and localization.

The present mechanism establishes category-level policy grounding. The resulting evidence can identify the predicted category, the localized computational evidence supporting that prediction, and the policy category with which the shared evidence representation is aligned. Section 3.4 next constrains which SAE features may constitute this evidence so that the resulting vocabulary remains feasible for inspection.

### 3.4 DR4. Evidence Inspectability: Task-Adaptive Sparse Evidence Vocabulary

Evidence inspectability requires the internal evidence available for review to be bounded, task-relevant, and organized so that its semantic structure can be inspected. Although a pretrained SAE produces sparse activations for individual segments, local sparsity does not by itself yield a bounded, task-specific evidence vocabulary, because different segments may activate different subsets of the full feature dictionary. SPACE therefore uses the global relevance gate $g$ introduced in Section 3 to learn task-specific relevance over the pretrained SAE feature space.

During training, SPACE encourages the gate to retain a limited number of features through an $L_1$ sparsity penalty:

$$L_{spar} = \|g\|_1 \tag{13}$$

The learned gate provides continuous feature relevance, whereas deployment requires a discrete evidence vocabulary. For each candidate threshold δ, SPACE retains

$$S_{\delta} = \left\{k \in \{1, ..., K\}: g_k > \delta\right\} \tag{14}$$

Each candidate vocabulary $S_{\delta}$ is evaluated on validation data using two deployment criteria: predictive performance $M_{val}(\delta)$, which should be maximized, and vocabulary size $|S_{\delta}|$, which should be minimized. A candidate is Pareto efficient when no alternative achieves equal or higher predictive performance with an equal or smaller vocabulary, with at least one strict improvement.

The final operating point is selected before test evaluation using the one-standard-error rule. Let $P$ denote the set of Pareto-efficient candidates, $M_{val}^{max}$ the highest validation performance among them, and $SE_{best}$ its standard error. SPACE selects

$$\delta^{\star} = argmin_{\delta \in P}\ |S_{\delta}| \quad subject\ to \quad M_{val}(\delta) \geq M_{val}^{max} - SE_{best} \tag{15}$$

This rule selects the smallest Pareto-efficient evidence vocabulary whose validation performance remains within one standard error of the best-performing candidate. The resulting $S_{\delta^{\star}}$ defines the deployment vocabulary used to expose feature-level evidence. Section 3.5 specifies how the training objectives governing prediction, localization, policy grounding, and sparsity are optimized over the shared gated representation.

### 3.5 Joint Learning of Governance-Ready Evidence

The mechanisms introduced in Sections 3.1–3.4 are optimized jointly over the shared gated representation. The central optimization challenge is to learn a shared feature-relevance structure under the combined requirements of prediction, evidence localization, policy grounding, and evidence inspectability. SPACE optimizes the evidence head, global gate, and policy projector simultaneously using the objective:

$$L = (L_{video} + \lambda_{seg} L_{seg}) + (\lambda_{local} L_{local} + \lambda_{cross} L_{cross}) + \lambda_{policy} L_{policy} + \lambda_{spar} L_{spar} \tag{16}$$

where $L_{video}$ and $L_{seg}$ govern predictive accuracy and segment supervision, while the remaining terms govern within-video localization, cross-video regularization, policy alignment, and gate sparsity, respectively.

The architecture and joint objective operationalize the four requirements over a shared gated representation. Computational traceability (DR1) is structurally enforced by the linear evidence architecture, whose decomposable scores are directly optimized by classification signals ( $L_{video}$, $L_{seg}$). Evidence localization (DR2) ($L_{local}$, $L_{cross}$) and policy grounding (DR3) ($L_{policy}$) pull the global gate $g$ to retain features that distinguish harmful content spans and align with policy concepts, while the sparsity objective $L_{spar}$ limits the density of the retained feature set (DR4). The shared gate $g$ is therefore the point at which prediction, localization, policy alignment, and sparsity jointly determine task-level feature relevance, producing the relevance structure that is subsequently converted into a discrete deployment vocabulary via Pareto selection.

### 3.6 Technical Contributions of the Research Design

At the system level, SPACE addresses the core design challenge established in Sections 1 and 2 by turning the structural interdependence among the four governance requirements into an integrated computational architecture. The evidence that supports the prediction is designed to remain simultaneously traceable to the computation, localizable within the video, interpretable against platform policy, and sufficiently bounded to permit human inspection. SPACE operationalizes this interdependence through a shared prediction-linked sparse evidence representation, allowing the mechanisms implementing the four requirements to act on a coherent evidentiary object.

At the algorithmic level, SPACE's first technical contribution: *governance-conditioned downstream sparse feature selection*, is designed around the joint requirements of *computational traceability* and *evidence inspectability*. The pretrained SAE exposes a large general-purpose feature dictionary whose full vocabulary is infeasible for human inspection, while computational traceability requires the retained evidence to remain directly connected to the prediction. SPACE introduces a global relevance gate $g$ learned downstream of SAE training through the joint objective in Eq. 16. This mechanism selects a bounded, task-specific evidence vocabulary $S_\tau$ from the pretrained SAE feature dictionary of dimension $D$, with $|S_\tau| \ll D$, from which localized decision evidence is instantiated through a highly sparse subset of active features ($k \ll |S_\tau|$), preserving exact linear traceability while limiting the evidence vocabulary to a scale feasible for human inspection.

The second technical contribution, *coupling within-video evidence ranking directly to decomposable decision evidence*, addresses the challenge created when fine-grained evidence localization and abstract policy grounding are jointly imposed on the same evidence representation. Localizing evidence on noisy video segments risks capturing irrelevant background signals unrelated to governance-relevant harm. SPACE addresses this by deriving segment harmfulness $h_{i,t}$ from the decision evidence scores in Eq. (7) and applying the within-video ranking objective in Eq. (8) directly to $h_{i,t}$. This coupling concentrates decision-critical evidence on sampled segments within ground-truth harmful time spans while preserving its exact mathematical connection to the final prediction within the shared representation used for policy alignment. Within-video ranking and L1-regularized gating are established algorithmic ingredients; SPACE's technical contribution lies in integrating these mechanisms over a shared, prediction-linked sparse evidence representation shaped jointly by governance requirements.

These algorithmic mechanisms operate jointly: governance-conditioned feature selection determines which internal SAE features form the inspectable vocabulary, while decision-linked within-video ranking determines where those features accumulate evidence under policy constraints. They transform a general-purpose SAE feature space into task-adapted evidence for governance-ready harmful-video detection.

# 4 EVALUATION

## 4.1 Data, Experimental Setup, and Evaluation Protocol

We constructed our dataset from TikTok, where toxic diet culture has been widely documented (Paul, 2021). Drawing on prior research on social media, body image, and disordered eating (Holland & Tiggemann, 2016), we selected hashtags representing prominent themes in this literature, such as #weightloss, #diet, #dietculture, #bodyimage, and #fasting. Using these hashtags, we retrieved approximately 11,000 candidate videos. We first screened the videos based on their associated hashtags to remove commercial content, followed by the exclusion of non-English videos. The remaining videos entered the annotation process.

Seven trained annotators labeled the videos using annotation guidelines grounded in TikTok's Mental and Behavioral Health policy (TikTok, 2024). After completing a 50-video qualification exercise, annotators assigned each video to Non-Toxic, Eating Disorders, or Body Anxiety Comparison. For harmful videos, they also identified the platform policy violated and marked the time spans containing harmful content. These temporal annotations provide training supervision for evidence localization and the ground truth for its evaluation. All videos entering the annotation stage were independently annotated by two annotators. Pairwise Cohen's kappa across annotator pairs averaged 0.83 and ranged from 0.80 to 0.91, indicating substantial to almost perfect agreement (Landis & Koch, 1977). We retained only videos for which the annotators agreed on the category label[2]. Online Appendix B provides the complete data-collection hashtag list, annotation protocol, operational category definitions, decision rules, and temporal-annotation procedure. This annotation protocol received our institutional ethics approval (Project No. PN-202500107; approved June 27, 2025), and all video and metadata were handled in accordance with standard research data-protection practices.

[2] We retained only videos for which the annotators agreed on the category label because supervised learning requires stable and consistent ground-truth signals during training. This criterion excludes some borderline cases with category disagreement, which should be considered when generalizing the findings to such cases. In total, 169 videos with discrepant category annotations were excluded, representing only 3.5% of the double-annotated sample.

After annotation and quality control, the corpus was divided into 3,343 training videos and 368 held-out test videos. We used creator account as the grouping variable, so all videos from a given content creator appear in only one partition. This design reduces the possibility that a model benefits from recurring creator-specific signals such as faces, backgrounds, or editing styles. Within the training partition, model selection and hyperparameter tuning used five-fold StratifiedGroupKFold with the same creator-level grouping rule. For each random seed, we selected the SPACE model with the highest validation score across folds and reported the mean and standard deviation of the selected models' held-out test performance across seeds. The held-out test set was not used for training, threshold selection, or hyperparameter tuning and was reserved for final evaluation. Following established frame-sampling practice in video moderation (Yang et al., 2026), SPACE represented each video by eight uniformly spaced visual observations; in SPACE, each observation served as one temporal unit for feature extraction and evidence scoring. Table 3 summarizes the category distribution across the training and held-out test sets.

**Table 3. Dataset description**

| Category | | Total | Training | Test |
|---|---|---|---|---|
| Non-Toxic | | 3,301 | 2,963 | 338 |
| Toxic | Eating Disorders | 281 | 267 | 14 |
| | Body Anxiety Comparison | 129 | 113 | 16 |

SPACE uses two pretrained components without further training. The pretrained LMM backbone is LLaVA-NeXT-LLaMA3-8B. We apply the publicly released sparse autoencoder developed for the same backbone and layer by K. Zhang et al. (2025). The autoencoder contains 131,072 latent features and uses a TopK constraint with K = 256. Encoding operates at the token level, so visual tokens from each sampled observation pass independently through the fixed sparse autoencoder. Within each sampled observation, token-level SAE activations are combined using max pooling to construct the segment-level feature vector used in Section 3. The resulting sparse activations are supplied to the shared global relevance gate and linear evidence head described in Section 3. The shared gated representation supports video classification, localized evidence scoring, and policy alignment. For policy alignment, a projection head maps the gated representation into the text-embedding space of a CLIP ViT-L/14 encoder (Cherti et al., 2023), where each policy category is represented by a fixed embedding of its policy description. Online Appendix A provides the remaining implementation and optimization details.

The evaluation protocol separates predictive performance from the additional functions required for governance. Because Non-Toxic videos dominate the corpus, the primary prediction metric is harmful-category Macro F1, computed as the unweighted mean of the F1 scores across the harmful categories so that each harmful category contributes equally. Precision and recall are reported alongside Macro F1, and the subgroup analysis reports class-level performance to show where errors occur[3]. All models are evaluated under the same creator-grouped data protocol.

### 4.2 Predictive Performance and Benchmark Comparison

We first examine SPACE's predictive performance for harmful-video detection. This test establishes whether SPACE can support its broader evidence-generation objectives while maintaining strong classification performance. We compare SPACE against four benchmark families under the evaluation protocol described in Section 4.1. (1) The first benchmark family represents conventional feature-based classification. Following the feature-extraction-plus-classification paradigm, we use an ImageNet-pretrained R(2+1)D network (Tran et al., 2018) as a frozen video feature extractor and apply L1-regularized logistic regression (Lasso), a support vector machine (SVM), k-nearest neighbors (KNN), and a random forest to the resulting video representations. (2) The second family consists of video representation models: X-CLIP (Ni et al., 2022), TimeSformer (Bertasius et al., 2021), VideoMAE V2 (L. Wang et al., 2023), and a visual-only adaptation of Xie et al. (2026) configured to our three-category task. These comparisons assess whether SPACE remains competitive when evaluated against models designed specifically for video representation learning. (3) The third benchmark family consists of contemporary video-capable LMMs: Qwen3-VL (Bai et al., 2025), Gemini 3.1 Pro (Google DeepMind, 2026), GPT-5.4 (OpenAI, 2026), InternVideo2.5 (Y. Wang et al., 2025), and VideoLLaMA3 (B. Zhang et al., 2025). We evaluate these LMMs using three prompting strategies: guideline prompting inspired by in-context learning, structured reasoning inspired by chain-of-thought prompting, and a combined strategy. Where model weights permit, Table 4 also reports supervised fine-tuning. The complete prompting and supervised fine-tuning protocols are provided in Online Appendix C. These models provide especially demanding benchmarks because they test whether powerful pretrained LMMs can solve the focal task through prompting or video-level fine-tuning. Table 4 therefore provides a system-level comparison of SPACE and the LMM benchmarks under their respective training protocols. (4) The fourth family comprises explainable and policy-aware video moderation methods: LEAF (Lang et al., 2026), MATCH (Li et al., 2026), and UNIVID (Yang et al., 2026). LEAF learns to generate explanations, MATCH verifies explanatory clues against retrieved video evidence, and UNIVID uses policy-aware video captions to support moderation. These benchmarks assess SPACE's predictive performance against closely related, task-specific moderation approaches.

[3] Unless otherwise stated, precision, recall, and F1 are computed separately for each harmful category and then macro-averaged. Consequently, harmful-category Macro F1 is not the harmonic mean of the reported macro-precision and macro-recall. Values reported as mean ± SD are calculated across ten randomly selected seeds; the same seed-selection rule is used for all such entries, and seed selection is independent of held-out test performance. Entries without ± report either the held-out test performance of the final configuration selected through hyperparameter tuning or, for prompted LMMs, identical results across repeated runs (SD = 0). For Table 4, metrics are computed with scikit-learn using zero_division = 0. When a harmful category is never predicted, its precision is set to 0 and remains included in the harmful-category macro average; a macro-precision of 0.5000 can therefore occur when one harmful category has precision 1.0000 and the other 0.

**Table 4. Predictive performance of SPACE and benchmark methods**

| Method | Description | Harmful-category Macro F1[d] | Precision | Recall |
|---|---|---|---|---|
| *Panel A: Feature-based benchmarks* | | | | |
| Lasso | R(2+1)D features | 0.3801 | 1.0000 | 0.2366 |
| SVM | R(2+1)D features | 0.2667 | 1.0000 | 0.1607 |
| KNN | R(2+1)D features | 0.4222 | 1.0000 | 0.2679 |
| Random Forest | R(2+1)D features | 0.2829 | 1.0000 | 0.1652 |
| *Panel B: Video representation-based benchmarks* | | | | |
| X-CLIP | — | 0.7430 ± 0.0717 | 1.0000 ± 0.0000 | 0.6022 ± 0.0889 |
| TimeSformer | — | 0.6555 ± 0.0835 | 1.0000 ± 0.0000 | 0.4988 ± 0.0911 |
| VideoMAE V2 | — | 0.6829 ± 0.0815 | 1.0000 ± 0.0000 | 0.5301 ± 0.0919 |
| Visual-only adaptation of Xie et al. (2026) | — | 0.6218 ± 0.0647 | 0.9689 ± 0.0198 | 0.4786 ± 0.0676 |
| *Panel C: Video-capable LMM benchmarks* [a,b] | | | | |
| Qwen3-VL | Prompt 1 | 0.5667 | 1.0000 | 0.5357 |
| | Prompt 2 | 0.3846 | 0.5000 | 0.3125 |
| | Prompt 3 | 0.4286 | 0.5000 | 0.3750 |
| | Fine-tuned | 0.7955 | 0.9757 | 0.6772 |
| InternVideo2.5 | Prompt 1 | 0.6089 | 1.0000 | 0.5402 |
| | Prompt 2 | 0.3631 | 1.0000 | 0.2277 |
| | Prompt 3 | 0.0667 | 0.5000 | 0.0357 |
| | Fine-tuned | 0.8007 | 0.9813 | 0.6786 |
| VideoLLaMA3 | Prompt 1 | 0.3600 | 0.5000 | 0.2812 |
| | Prompt 2 | 0.4074 | 0.5000 | 0.3438 |
| | Prompt 3 | 0.2381 | 0.5000 | 0.1562 |
| | Fine-tuned | 0.7858 | 0.9859 | 0.6571 |
| GPT-5.4 | Prompt 1 | 0.7333 | 0.9643 | 0.6205 |
| | Prompt 2 | 0.7667 | 1.0000 | 0.6518 |
| | Prompt 3 | 0.7333 | 0.9643 | 0.6205 |
| Gemini 3.1 Pro (matched-input benchmark: 8 sampled observations) | Prompt 1 | 0.7407 | 1.0000 | 0.5938 |
| | Prompt 2 | 0.7816 | 1.0000 | 0.6562 |
| | Prompt 3 | 0.6970 | 1.0000 | 0.5357 |
| Gemini 3.1 Pro (full video) | Prompt 1 | 0.8119 | 1.0000 | 0.6920 |
| | Prompt 2 | 0.7483 | 1.0000 | 0.6205 |
| | Prompt 3 | 0.7513 | 1.0000 | 0.6027 |
| *Panel D: Explainable and policy-aware video moderation benchmarks* [c] | | | | |
| LEAF | Two-stage reasoning and explanation distillation | 0.7945 ± 0.0585 | 0.9221 ± 0.0490 | 0.7045 ± 0.0671 |
| MATCH | Clue proposal, evidence retrieval, and verification | 0.6958 ± 0.0349 | 0.9844 ± 0.0251 | 0.5424 ± 0.0333 |
| UNIVID-Lite | Policy-aware captioning and moderation fine-tuning | 0.5993 ± 0.1134 | 1.0000 ± 0.0000 | 0.4384 ± 0.1134 |
| UNIVID-RAG | Policy-aware captioning and retrieval-augmented moderation | 0.5876 ± 0.0712 | 1.0000 ± 0.0000 | 0.4214 ± 0.0759 |
| **SPACE (Ours)** | — | **0.8507 ± 0.0103** | **0.9958 ± 0.0132** | **0.7469 ± 0.0172** |

**Note:** [a] Unless otherwise noted, the LMM benchmarks receive the same eight uniformly sampled visual observations used by SPACE. Gemini 3.1 Pro is additionally evaluated with full-video input.

[b] Panel C. Prompt 1 is inspired by in-context learning and supplies the annotation guidelines in the model context (without demonstration examples). Prompt 2 is inspired by chain-of-thought prompting and guides the model through predefined diagnostic questions. Prompt 3 combines the annotation guidelines with the same structured reasoning procedure. Fine-tuned results are reported where model weights permit. The complete prompts, API model and inference settings, and supervised fine-tuning protocols are provided in Online Appendix C.

[c] Panel D: Implementations follow LEAF's released code and the published procedures for MATCH and UNIVID's Moderation Actor, with task-specific adaptations to a shared LLaVA-NeXT-LLaMA3-8B backbone, visual-only inputs, and the three-category task. Online Appendix D documents these adaptations.

[d] Given the limited harmful-category test counts, between-method performance differences are interpreted descriptively.

Table 4 shows that SPACE achieves the highest harmful-category Macro F1 among the evaluated methods, at 0.8507. It also achieves the highest recall at 0.7469, with a reported precision of 0.9958. Among the feature-based benchmarks, KNN performs best with a Macro F1 of 0.4222. One possible explanation for this difference is that these methods classify a video-level representation, whereas SPACE learns evidence at the segment level and allows the strongest segment to contribute directly to the video-level decision. This design is particularly relevant when harmful evidence occupies only a small portion of an otherwise benign video.

Among the video representation benchmarks, X-CLIP achieves the strongest performance, with a Macro F1 of 0.7430. These models provide substantially richer spatiotemporal representations than the feature-based benchmarks, but their training and prediction remain centered on the video as the unit of classification. SPACE additionally uses harmful time-span annotations to distinguish harmful from benign segments within the same video and learns a task-adaptive gate that emphasizes sparse features useful for the focal detection task. These task-specific learning signals may help explain the additional performance gain.

The strongest results among the video-capable LMMs are closer to SPACE than those of the other benchmark families. Because Gemini 3.1 Pro was the strongest LMM available at the time of our evaluation, we tested it using both the matched eight-observation input and the unrestricted full-video input (H. He et al., 2026). Under the matched input, Gemini 3.1 Pro achieves its highest Macro F1 of 0.7816 under Prompt 2, while fine-tuned InternVideo2.5 reaches 0.8007, Qwen3-VL reaches 0.7955, and VideoLLaMA3 reaches 0.7858. With full-video input, Gemini 3.1 Pro achieves a Macro F1 of 0.8119 under Prompt 1, compared with 0.8507 for SPACE. Notably, SPACE achieves this performance with the 8B LMM backbone and pretrained SAE feature space held fixed; only the task-specific gate, evidence head, and policy projector are trained. Prompt-based LMMs benefit from strong pretrained multimodal semantic capabilities but do not update their parameters for the focal task, whereas fine-tuned LMMs adapt to the training data using video-level category labels. SPACE similarly starts from pretrained multimodal representations but incorporates segment-level supervision, within-video contrastive learning, and task-adaptive feature selection. The closer alignment of these learning objectives with the structure of harmful short-video detection may help explain the observed performance differences between SPACE and the LMM benchmarks.

Among the adapted explainable and policy-aware moderation benchmarks, LEAF achieves the highest reported harmful-category Macro F1 at 0.7945, followed by MATCH at 0.6958, UNIVID-Lite at 0.5993, and UNIVID-RAG at 0.5876. SPACE exceeds the strongest benchmark in this family by 0.0562. The performance differences are particularly evident in recall, which is lower for all four benchmarks than for SPACE. One plausible explanation is the closer alignment between SPACE's training signals and the focal prediction task. SPACE applies segment-level supervision and within-video ranking directly to the category evidence scores used for video prediction, which may improve sensitivity when harmful signals appear briefly within otherwise benign videos.

### 4.3 Evaluation of Governance-Ready Evidence

We next examine whether SPACE generates *governance-ready evidence* that is computationally traceable, localized, policy-grounded, and inspectable. Our review of prior work identifies no approach that jointly supports these four capabilities within a decision-linked evidence structure. We therefore adopt three complementary evaluation strategies: an integrated evidence illustration, capability-specific benchmark comparisons, and controlled design comparisons.

**Governance-ready evidence illustration.** Figure 3 presents evidence reports for videos classified as Body Anxiety Comparison and Eating Disorders. In the Body Anxiety Comparison example, the report identifies torso-focused observations and associated SAE features with candidate glosses such as "weight" and "body fat." In the Eating Disorders example, it identifies observations of the eating episode and features associated with "pizza," "eating," and "consume food." Each report also identifies the corresponding policy category.

Across these examples, the timestamped observations locate the relevant content (DR2), and the feature contributions expose the computational support for the prediction (DR1). These contributions are the feature-level terms used to compute the segment-category scores defined in Section 3.1. The policy category provides the external standard for evaluating the content (DR3), while the focused feature list and semantic glosses make the internal evidence available for inspection (DR4). This structure allows a moderator to examine localized, computationally traceable, policy-grounded, and inspectable evidence supporting the model's classification of a video as harmful. These linked forms of evidence address the four governance questions in Table 1: *where* harmful content occurs, *what computational evidence* supports its detection, *which policy* applies, and *how* the evidence can be inspected.

**Figure 3. Example SPACE Evidence Report Linking Localized Harmful Evidence, Sparse Feature Contributions, and Policy Category**

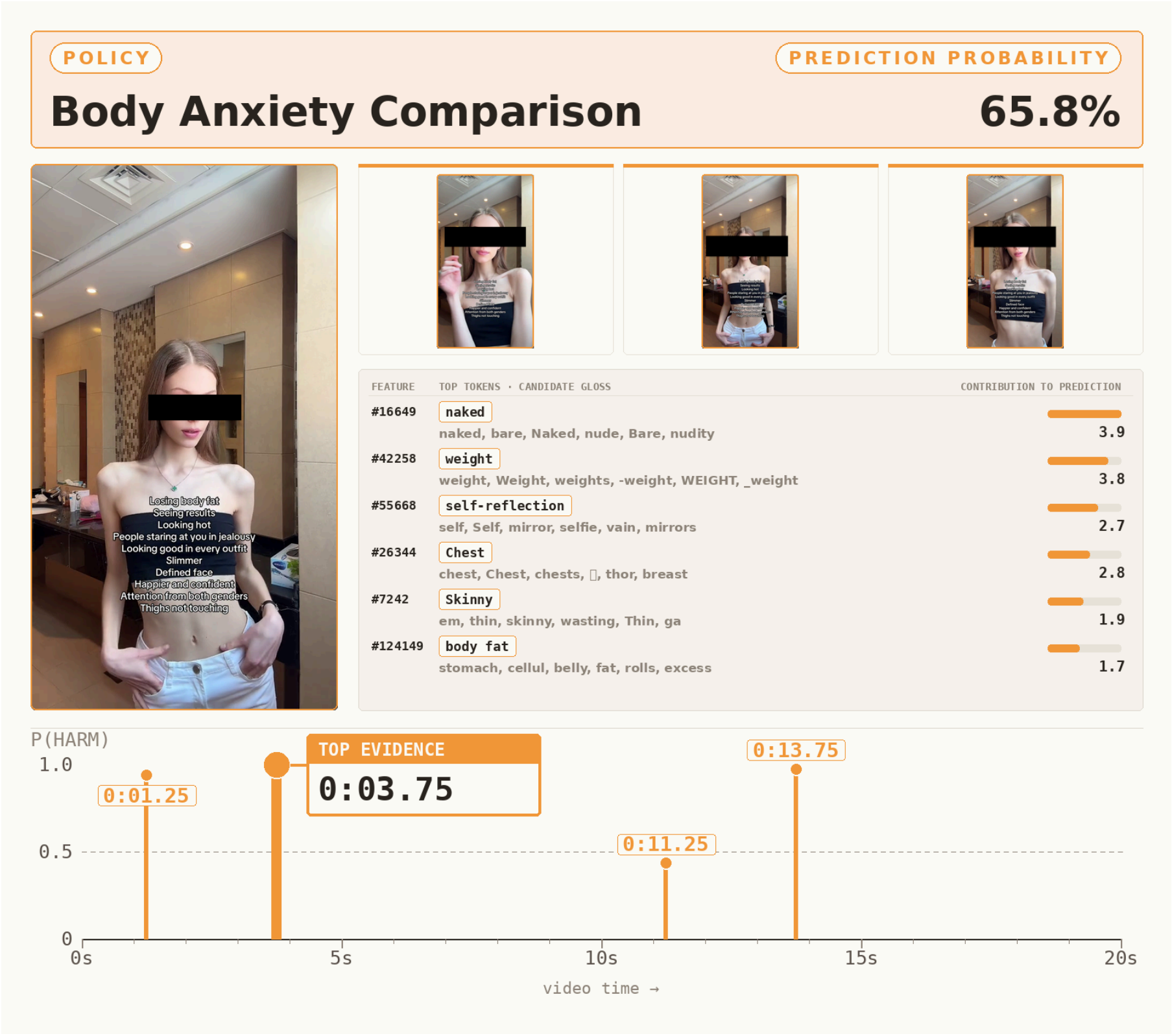

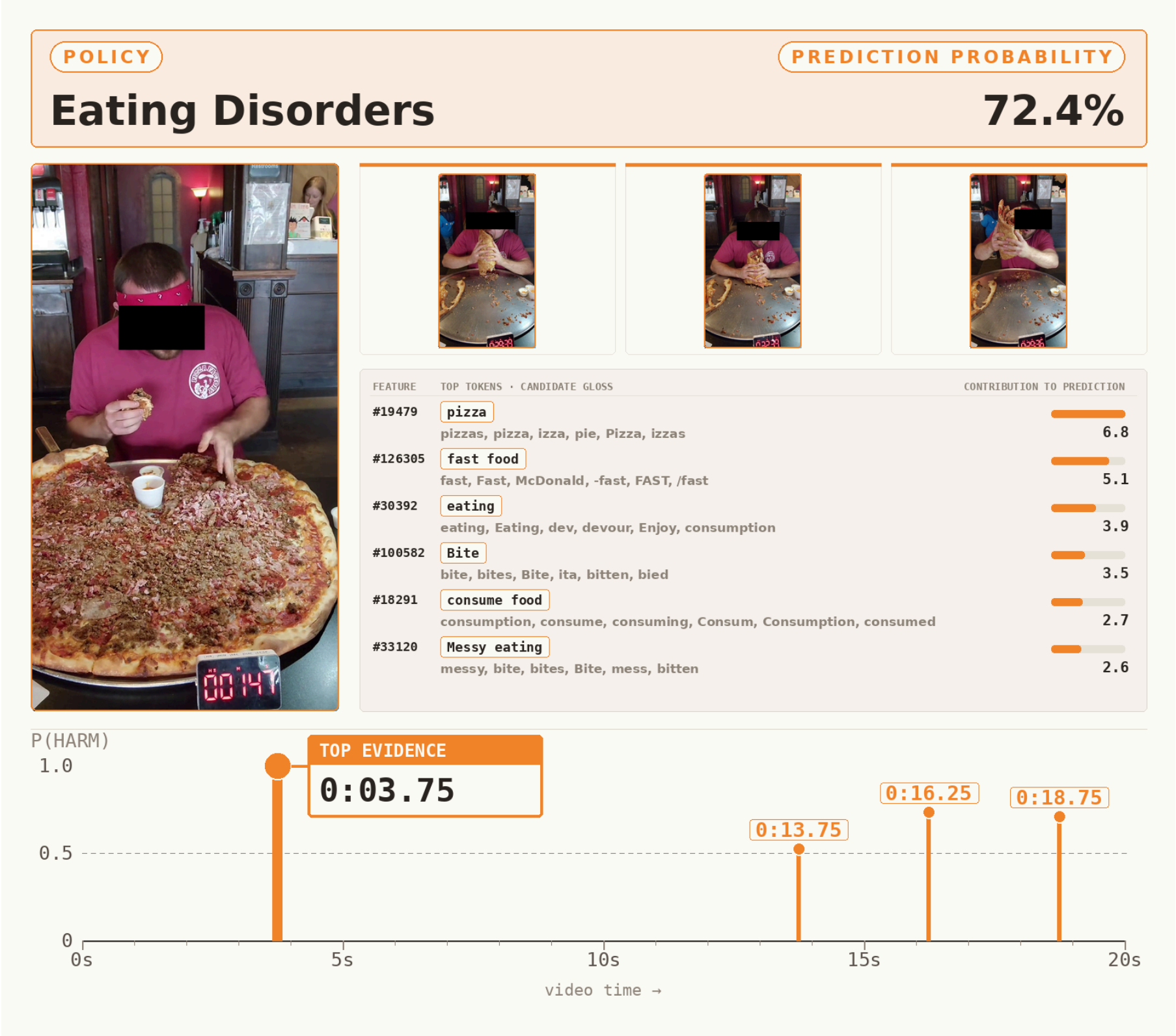

Note. Feature glosses indicate candidate semantic associations of contributing SAE features; their contributions to harmful-content predictions depend on gated activations and learned category-specific weights. Policy grounding is at the category level.

The following analyses examine the quality of the governance-ready evidence.

**Evidence localization (DR2).** Table 5 examines the contribution of SPACE's localization mechanisms in Panel A and compares its localization and classification performance with adapted MATCH in Panel B. Because annotated harmful time spans cover multiple sampled observations in many harmful videos, Panel A includes a random-frame baseline for Top-1 Evidence Hit Rate to provide a reference for evaluating localization performance.

**Table 5. Evidence Localization: Design Comparisons and Benchmark Evaluation**

| Method | Harmful-category Macro F1 | Top-1 Evidence Hit Rate [a] |
|---|---|---|
| *Panel A. Localization design* | | |
| SPACE (Ours) | 0.8507 ± 0.0103 | 0.8800 ± 0.0358 |
| w/o within-video ranking | 0.7542 ± 0.0563 | 0.8367 ± 0.0693 |
| w/o segment supervision | 0.7640 ± 0.0670 | 0.8767 ± 0.0545 |
| w/o within-video ranking & segment supervision | 0.7204 ± 0.0638 | 0.8033 ± 0.0532 |
| Random-frame baseline [b] | – | 0.8458 |
| *Panel B. Localization-capable benchmark* | | |
| SPACE (Ours) | 0.8507 ± 0.0103 | 0.8800 ± 0.0358 |
| MATCH | 0.6958 ± 0.0349 | 0.7667 ± 0.0534 |

**Note.** [a] Top-1 Evidence Hit Rate measures the proportion of harmful test videos for which the timestamp of the top-ranked sampled observation falls within an annotated harmful time span. SPACE and its variants rank sampled observations using the harmfulness score in Eq. (7).
[b] For each harmful test video, the random-frame baseline is the proportion of its eight sampled observations whose timestamps fall within annotated harmful time spans; the reported baseline averages this proportion across harmful test videos.

Table 5 (Panel A) shows that the random-frame baseline is relatively high at 0.8458, reflecting the substantial temporal coverage of annotated harmful content in the test videos. SPACE nevertheless achieves a higher Top-1 Evidence Hit Rate of 0.8800, indicating that its strongest decision evidence falls within annotated harmful time spans more consistently than random selection. Removing within-video ranking (Eq. 8) reduces the hit rate to 0.8367, below the random-frame baseline, and lowers harmful-category Macro F1 from 0.8507 to 0.7542. Removing segment supervision leaves the hit rate nearly unchanged at 0.8767 but lowers Macro F1 to 0.7640. Removing both reduces the hit rate further to 0.8033 and Macro F1 to 0.7204. This pattern is consistent with within-video ranking contributing to the temporal placement of decision evidence, while segment supervision primarily strengthens category discrimination. Their combined use achieves the highest mean localization and classification scores among the evaluated SPACE variants.

Table 5 (Panel B) shows that SPACE exceeds adapted MATCH in both Top-1 Evidence Hit Rate (0.8800 versus 0.7667) and harmful-category Macro F1 (0.8507 versus 0.6958). The localization advantage of 0.1133 indicates that SPACE is more likely than the benchmark to select evidence within the harmful time span.

In addition, SPACE derives its localization score directly from the decomposable category scores used for prediction, thereby linking evidence localization (DR2) to computational traceability (DR1).

**Policy grounding (DR3).** Table 6 addresses two questions: how the policy-alignment design affects predictive performance and policy mapping, and how SPACE compares with existing policy-capable methods. Panel A compares alternative alignment designs within SPACE; Panel B compares SPACE with the adapted UNIVID-Lite and UNIVID-RAG benchmarks. To evaluate the contribution of the policy-alignment objective and the choice of alignment representation, we compare SPACE with two controls: (1) the segment-level condition aligns the same policy descriptions with harmful sampled observations, whereas SPACE aligns the pooled video representation; (2) the no-policy condition removes the policy-alignment loss while keeping the rest of the model unchanged.

**Table 6. Policy Grounding: Design Alternatives and Capability Benchmarks**

| Method / configuration | Harmful-category Macro F1 | Policy Mapping Accuracy [b] |
|---|---|---|
| *Panel A. Alignment design* [a] | | |
| SPACE: video-pooled alignment [a] | 0.8507 ± 0.0103 | 0.8700 ± 0.0761 |
| Harmful-segment alignment | 0.7306 ± 0.0562 | 0.9033 ± 0.0555 |
| No policy-alignment loss | 0.7465 ± 0.0692 | — |
| *Panel B. Policy-capable benchmarks* | | |
| SPACE (Ours) | 0.8507 ± 0.0103 | 0.8700 ± 0.0761 |
| UNIVID-Lite | 0.5993 ± 0.1134 | 0.4400 ± 0.1116 |
| UNIVID-RAG | 0.5876 ± 0.0712 | 0.4200 ± 0.0730 |

Note. [a] SPACE aligns the pooled video evidence representation with the policy description corresponding to each content category. Harmful-segment alignment uses evidence from harmful segments or frames, paired with the corresponding category-level policy description. No policy alignment sets the policy-alignment loss weight to zero, so no evidence representation is aligned with policy text. [b] Policy Mapping Accuracy is the proportion of harmful test videos for which the policy description with the highest similarity matches the ground-truth policy category. A dash indicates that policy mapping was not evaluated.

SPACE's video-pooled alignment achieves a Macro F1 of 0.8507 and Policy Mapping Accuracy of 0.8700. Segment-level alignment increases policy accuracy to 0.9033 but reduces Macro F1 to 0.7306. This comparison reveals an empirical trade-off: the adopted configuration retains substantially higher predictive performance while achieving strong policy-mapping accuracy. Removing policy alignment reduces Macro F1 from 0.8507 to 0.7465, demonstrating the contribution of the policy-alignment objective to predictive performance under the evaluated training protocol.

Policy Mapping Accuracy evaluates whether the learned evidence identifies the applicable policy for a harmful video. It is calculated as the proportion of harmful test videos for which the policy description with the highest similarity corresponds to the ground-truth policy category. As shown in Table 6 (Panel B), SPACE achieves a Policy Mapping Accuracy of 0.8700, compared with 0.4400 for UNIVID-Lite and 0.4200 for UNIVID-RAG. These results support the policy-grounding requirement (DR3): SPACE's evidence representation identifies the policy category applicable to the harmful video. Because policy alignment operates on a pooled representation derived from the shared gated representation used for prediction and localization, this capability connects the computational evidence (DR1) supporting a judgment to its governing policy (DR3)[4].

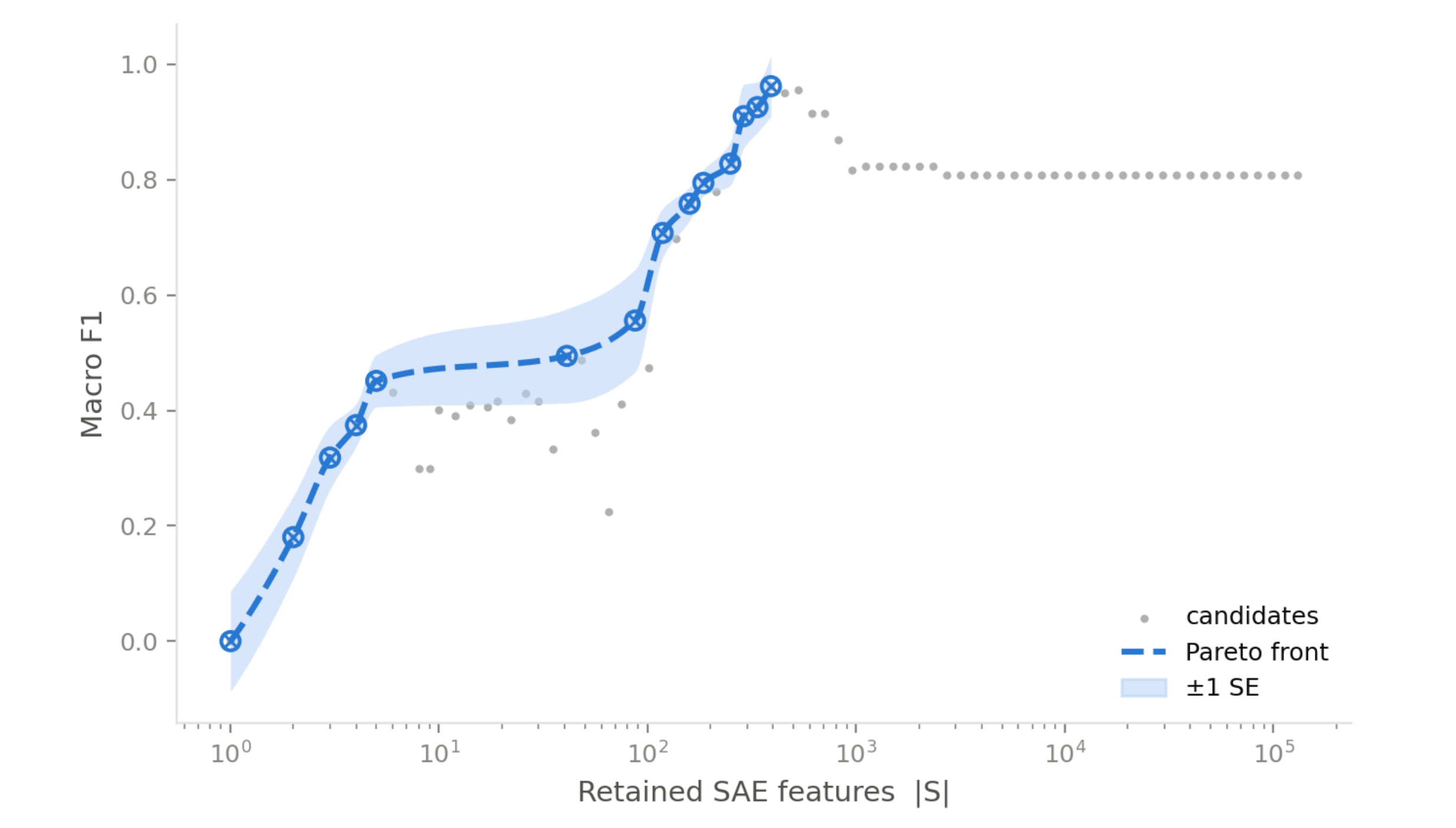

**Figure 4. Pareto Frontier of Predictive Performance and Feature Parsimony**

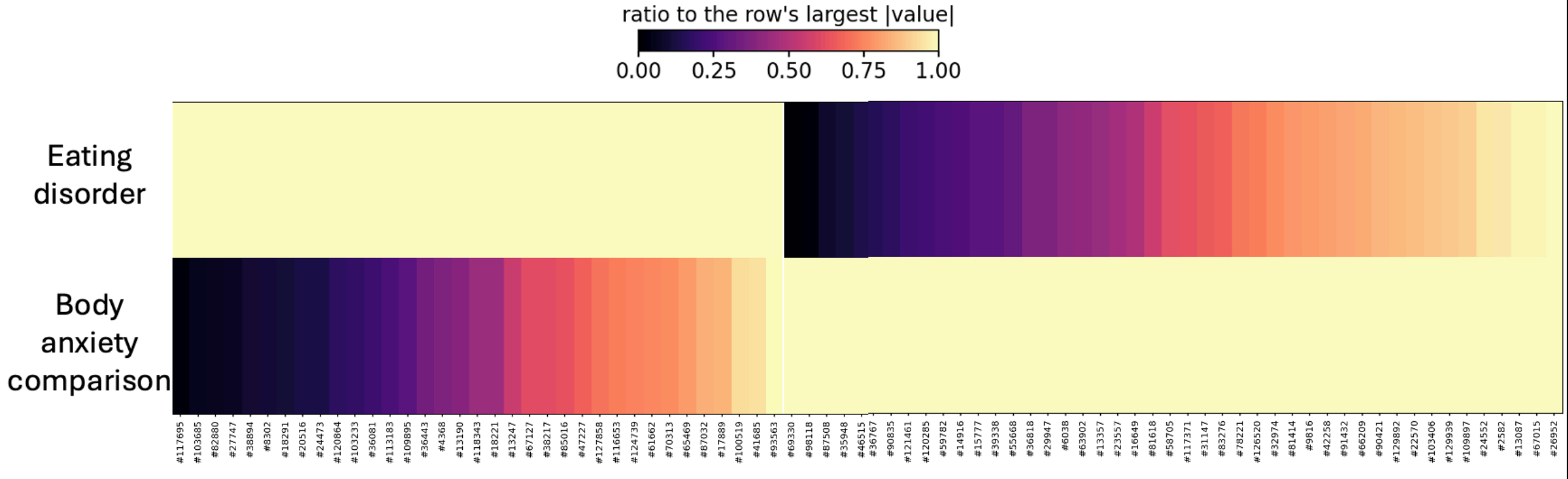

**Figure 5. Category Structure of Selected SAE Features (Mean Activation on Harmful Segments)**

**Evidence inspectability (DR4).** Next, we examine the size and semantic structure of the selected evidence vocabulary. The analyses assess predictive performance across vocabulary sizes, category-specific activation patterns, and consistency among top-activating observations.

[4] Policy grounding refers to category-level correspondence between the learned evidence representation and the applicable policy description, rather than the contribution of individual policy-text semantics.

The task-adaptive selection mechanism combines learned feature relevance with Pareto-based selection to derive a compact evidence feature set from the pretrained SAE dictionary of 131,072 features for harmful-video detection. Figure 4 evaluates this design through the predictive performance and retained feature count of the candidate vocabularies. The Pareto frontier identifies configurations offering efficient performance–parsimony trade-offs, and the one-standard-error rule specified in Section 3.4 selects the final operating point before test evaluation. This procedure makes feature selection an explicit performance–parsimony design decision, determining how many features from the pretrained SAE feature space are retained in the task-specific evidence vocabulary.

The Pareto frontier also reveals the empirical structure of this trade-off. Extremely small feature sets provide insufficient predictive utility, while performance improves as task-relevant features are retained and eventually reaches a region in which adding substantially more features yields little additional benefit. The selected operating point identifies a compact feature set whose validation performance remains within the specified tolerance of the best-performing candidate. These results support selecting a task-specific evidence vocabulary from the pretrained SAE feature dictionary, with its size determined jointly by predictive utility and parsimony. This analysis addresses the vocabulary-size component of evidence inspectability (DR4), while the retained features remain linked to harmful-video predictions through the decomposable evidence architecture required for computational traceability (DR1).

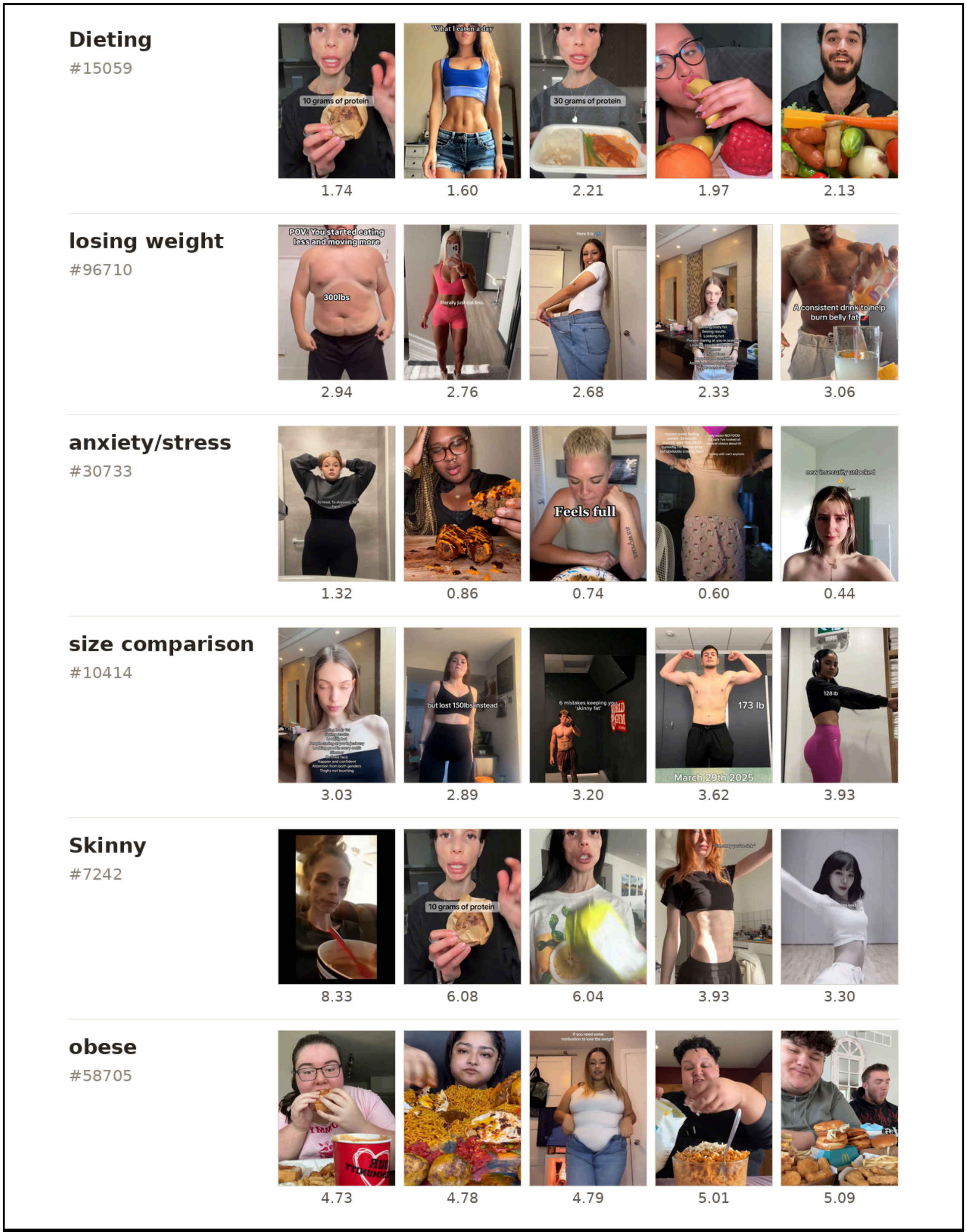

**Figure 6. Top-Activating Video Observations for Selected SAE Features**

Note: The number below each frame indicates the strength of the corresponding SAE feature response for that frame.

The next question is whether the retained features are substantively related to the harmful-content task. Figure 5 provides an illustrative example through a feature–category heatmap. The heatmap reveals how individual features concentrate their activations in particular harmful-content categories or respond more broadly across them. Such structure is important because task-adaptive selection should do more than retain a small number of latent dimensions; the retained vocabulary should reflect distinctions relevant to the downstream governance problem. The observed category-concentrated patterns therefore provide descriptive evidence that the selected features are systematically associated with harmful-content categories, supporting the task-relevance component of evidence inspectability (DR4).

Figure 6 examines the selected vocabulary from a complementary, instance-level perspective. For representative retained features, we retrieve the video observations that produce the strongest feature responses and display the top examples together. The numbers below the frames indicate the strength of each feature's response to the corresponding observation. This analysis evaluates whether an individual feature repeatedly responds to substantively similar visual content across different videos. Feature #96710, for example, repeatedly responds to weight-loss and body-transformation cues, while feature #10414 responds to body-size comparison imagery. These recurring visual associations make the semantic structure of the displayed features directly inspectable. Together with the feature-level score decomposition defined in Section 3.1, the selected vocabulary allows the contributions of these inspected features to be traced directly to harmful-video predictions (DR1).

### 4.4 Ablation and Performance Robustness

We next examine whether SPACE's performance depends on the two core technical contributions, whether governance-conditioned feature selection identifies more task-relevant features than random and activation-based matched-size selection, and whether performance is consistent across harmful-content categories.

**Ablation of the two core technical contributions.** SPACE contains two testable technical contributions. Technical contribution 1 is governance-conditioned downstream sparse feature selection, which selects a compact, task-adapted evidence feature set from the pretrained SAE feature space. Technical contribution 2 introduces decision-linked within-video evidence ranking, which prioritizes decomposable evidence used for prediction and concentrates it on the ground-truth harmful video segments. Table 7 summarizes the evaluation of the two core technical contributions and other design components: Panel A reports the matched-size substitution test for Technical Contribution 1 and the ablation of Technical Contribution 2, while Panel B summarizes other evaluated design components. Tables 5 and 6 provide the corresponding mechanism-specific localization and policy-grounding results.

Replacing the task-adapted vocabulary with a matched-size random vocabulary under the fixed-parameter substitution test reported in Table 8 reduces harmful-category Macro F1 from 0.8507 to 0.5091. Removing decision-linked within-video ranking reduces Macro F1 to 0.7542. These results indicate that both the identity of the selected features and the direct coupling of localization to decision evidence contribute materially to SPACE's performance.

**Table 7. Ablation Analyses of SPACE Design Components**

| Configuration | Harmful-category Macro F1 | Precision | Recall |
|---|---|---|---|
| *Panel A. Core technical contributions* | | | |
| SPACE (Full) | 0.8507 ± 0.0103 | 0.9958 ± 0.0132 | 0.7469 ± 0.0172 |
| w/o Technical Contribution 1: *Governance-conditioned feature selection* [a] | 0.5091 ± 0.2420 | 0.7150 ± 0.3233 | 0.4275 ± 0.1769 |
| w/o Technical Contribution 2: *Decision-linked within-video evidence ranking* [b] | 0.7542 ± 0.0563 | 0.9958 ± 0.0132 | 0.6179 ± 0.0716 |
| *Panel B. Other design component ablations* | | | |
| SPACE (Full) | 0.8507 ± 0.0103 | 0.9958 ± 0.0132 | 0.7469 ± 0.0172 |
| w/o Segment Supervision [c] | 0.7640 ± 0.0670 | 1.0000 ± 0.0000 | 0.6272 ± 0.0868 |
| w/o Policy-Alignment Loss [d] | 0.7465 ± 0.0692 | 0.9962 ± 0.0122 | 0.6080 ± 0.0865 |
| Pretrained LMM Backbone Control [e] | 0.6955 ± 0.0579 | 1.0000 ± 0.0000 | 0.5402 ± 0.0676 |

Note: All ablation variants use the hyperparameter settings tuned for the full SPACE model, with only the specified component removed. [a] Technical Contribution 1 is evaluated using the matched-size random-N substitution test under fixed SPACE parameters reported in Table 8. [b] Technical Contribution 2 is ablated by removing the within-video ranking objective in Eq. (8) while retaining segment supervision and all other training objectives. [c] The segment-level classification term is removed while within-video ranking and the other training objectives are retained. [d] The policy-alignment loss coefficient is set to zero while the remaining model is unchanged. [e] The backbone control uses the same pretrained LLaVA-NeXT-LLaMA3-8B backbone as SPACE, with the backbone fixed and only a sigmoid classification head trained on the backbone representation; it does not use the SAE-based evidence

mechanisms introduced by SPACE.

**Task relevance of the selected evidence vocabulary at the Pareto operating point**. We next assess the task relevance of the selected evidence vocabulary through matched-size feature-subset comparisons. Holding the vocabulary size fixed at *N*, we compare the SPACE-selected vocabulary with activation-based Top-*N* and random-*N* feature subsets. For these matched-size comparisons, the trained SPACE parameters are held fixed; only the retained feature indices are changed, while the learned gate values and evidence-head parameters are preserved and no retraining is performed. This comparison therefore tests whether the SPACE-selected features are more task-relevant than equally sized feature subsets obtained through random selection or an existing activation-based selection approach.

**Table 8. Predictive Performance of Matched-Size Feature Subsets Under Fixed SPACE Parameters**

| Feature-selection strategy | Harmful-category Macro F1 | Precision | Recall |
|---|---|---|---|
| SPACE: Governance-conditioned gate with Pareto 1-SE selection (*N* features) | 0.8507 ± 0.0103 | 0.9958 ± 0.0132 | 0.7469 ± 0.0172 |
| Activation-based top-*N* feature selection (Gallifant et al., 2025) | 0.5643 ± 0.3989 | 0.5650 ± 0.4139 | 0.5795 ± 0.3856 |
| Random-*N* feature selection | 0.5091 ± 0.2420 | 0.7150 ± 0.3233 | 0.4275 ± 0.1769 |

Note: Reported SPACE results use 10 randomly selected training seeds. For each seed-specific SPACE run, *N* is the number of features selected by SPACE at the Pareto 1-SE operating point, and each matched-size alternative retains that same *N*. For the random-*N* condition, five independent random feature subsets of that seed-specific size *N* are drawn. These five feature draws are distinct from the 10 training seeds. Across the 10 SPACE runs, the median selected *N* is 869.5. Activation-based Top-N selection adapts prior Top-N filtering of SAE activations for downstream classification (Gallifant et al., 2025).

Table 8 shows that replacing the SPACE-selected vocabulary with a matched-size activation-based Top-N subset reduces harmful-category Macro F1 from 0.8507 to 0.5643, while random-N substitution yields 0.5091. These results indicate that, for the trained SPACE predictor, predictive performance depends on the identity of the retained features and not only on vocabulary size.

**Category-level subgroup analysis.** Harmful-category Macro F1 gives equal weight to the two harmful categories, but it can still conceal category-level heterogeneity. Table 9 therefore reports class-specific precision, recall, and F1. SPACE performs strongly for both Eating

Disorders, with an F1 of 0.8868, and Body Anxiety Comparison, with an F1 of 0.8145. Precision remains high in both categories, whereas recall is lower for Body Anxiety Comparison than for Eating Disorders, 0.6937 versus 0.8000. The remaining errors are therefore concentrated primarily in missed Body Anxiety Comparison videos. The lower recall suggests that the Body Anxiety Comparison category, which has fewer samples, is more difficult to detect.

**Table 9. Performance by Harmful-Content Category**

| Category | F1 | Precision | Recall |
|---|---|---|---|
| Aggregate across harmful categories | 0.8507 ± 0.0103 | 0.9958 ± 0.0132 | 0.7469 ± 0.0172 |
| Eating Disorders | 0.8868 ± 0.0513 | 1.0000 ± 0.0000 | 0.8000 ± 0.0811 |
| Body Anxiety Comparison | 0.8145 ± 0.0496 | 0.9917 ± 0.0264 | 0.6937 ± 0.0688 |

Note. The aggregate row reports the unweighted macro average across the two harmful categories; the category rows report class-specific metrics.

### 4.5 Stakeholder Evaluation of Perceived Governance Utility

We next examine the perceived governance utility of the resulting evidence report for the two stakeholder groups directly involved in platform moderation: content creators who receive moderation decisions and platform professionals who review and operationalize those decisions (Table 10). We conducted a stakeholder panel study with 76 content creators and 11 platform moderation professionals. The study used six real videos from our dataset, each correctly classified as harmful by SPACE and accompanied by a SPACE-generated report. Each report presented the predicted harmful-content category, the temporal observations carrying the strongest evidence, the contributing sparse features, and the corresponding policy information. Participants rated statements on a seven-point Likert scale from 1 (strongly disagree) to 7 (strongly agree), with 4 as the midpoint. Because the items represented distinct dimensions of governance utility, we analyze each item separately. We also report the overall mean as a descriptive summary. Stakeholder study materials are provided in Online Appendix E.

For content creators, the results indicate that the report is useful primarily because it makes a moderation decision easier to understand and act on. Across the six items, the overall mean was 6.10 (SD = 0.79), and 98% of ratings were 5 or higher. The highest-rated item was greater transparency than a generic warning or deletion notice (M = 6.37, SD = 0.80), followed by support for deciding how to revise a video before posting (M = 6.17, SD = 0.76) and identifying the harmful segment (M = 6.14, SD = 0.89). Creators also rated the report highly for understanding why a video was harmful (M = 5.99), relating the video to platform policy (M = 5.93), and anticipating its effect on sensitive viewers (M = 5.97). These results suggest that the value of the report to content creators extends beyond receiving a harmful-content label. The localized evidence and policy information provide a more specific basis for understanding what triggered the decision and what could be changed. Free-text responses were consistent with this use: creators asked for more specific identification of harmful words or cues, pre-publication checking or rechecking, broader analysis of the full video, and guidance on revisions that could remove a restriction.

The platform panel provides complementary evidence about organizational use. The 11 participants worked in content-moderation and live-stream operations roles and had a mean of 4.4 years of experience, with a range of 2 to 10 years. Across nine items, the overall mean was 6.04 (SD = 0.77), and 99% of ratings were 5 or higher. The highest-rated item was helping moderators focus on the key frames needed for a judgment (M = 6.27, SD = 0.79). Reducing moderator workload and supporting the recording, auditing, or explanation of decisions were both rated 6.18, while providing a sufficient basis for a policy decision and serving as feedback to creators were both rated 6.09. The remaining items, including use as moderator training material, support for creator content management, and personalization of the visualization based on the video creator and the platform's strategy, also received means above 5.7. Given the small platform panel, we focus on descriptive magnitudes and agreement rates. The responses indicate that participants perceived the report as useful for several parts of the moderation workflow: directing attention to the most relevant video evidence, connecting that evidence to a policy basis, preserving information that can be reviewed later, and communicating a decision back to the creator. Platform professionals also suggested concrete deployment features, including direct integration into the moderation backend, one-click navigation to risky segments, role-specific report views, graded-action controls, and exportable records for audit and compliance activities.

| Table 10. Stakeholder Evaluation of the SPACE Evidence Report | | |
|---|---|---|
| **Stakeholder and dimension** | **Mean ± SD** | **Agreement** |
| *Panel A: Content Creators (n = 76)* | | |
| Understanding why a video is harmful | 5.99 ± 0.66 | 100% |
| Identifying the harmful segment | 6.14 ± 0.89 | 96% |
| Relating the video to platform policy | 5.93 ± 0.75 | 100% |
| Deciding how to revise before posting | 6.17 ± 0.76 | 99% |
| Transparency over a generic warning or deletion notice | 6.37 ± 0.80 | 97% |
| Anticipating impact on sensitive viewers | 5.97 ± 0.78 | 97% |
| Overall | 6.10 ± 0.79 | 98% |
| *Panel B: Platform Moderation Professionals (n = 11)* | | |
| Helping creators understand why a video was blocked | 5.82 ± 0.75 | 100% |
| Reducing moderator workload | 6.18 ± 0.75 | 100% |
| Focusing on the key frames for a judgment | 6.27 ± 0.79 | 100% |
| Providing a sufficient basis for a policy decision | 6.09 ± 0.83 | 100% |
| Recording, auditing, or explaining decisions | 6.18 ± 0.87 | 100% |
| Serving as moderator training material | 5.91 ± 0.70 | 100% |
| Serving as feedback to creators | 6.09 ± 0.70 | 100% |
| Helping creators manage and optimize content | 5.73 ± 0.90 | 91% |
| Personalizing the visualization based on creator and platform strategy | 6.09 ± 0.70 | 100% |
| Overall | 6.04 ± 0.77 | 99% |

Note: Agreement denotes the percentage of ratings at 5 or above on the seven-point scale.

# 5 DISCUSSION AND CONCLUSION

This study developed and evaluated SPACE to address a central limitation of existing harmful-video detection: accurate classification alone provides limited support for platform governance when moderators also need to know what evidence supports a decision and how that evidence relates to policy. SPACE contributes a unified computational design that makes computational traceability, evidence localization, policy grounding, and evidence inspectability properties of the same predictive artifact built on a shared gated representation. The evaluation shows that this design maintains strong predictive adequacy while producing localized, computationally traceable, policy-grounded, and inspectable evidence that stakeholders perceive as useful for moderation review. These findings answer the study's research question by showing how governance requirements can shape both the architecture and evaluation of harmful-video detection.

## 5.1 Technical Contributions and Design Principles

SPACE’s system-level contribution is to instantiate four structurally interdependent governance requirements within a coherent prediction-linked evidence architecture. This challenge cannot be reduced to adding four independent capabilities because the validity of each form of evidence depends on its relationship to the others. SPACE addresses this interdependence through a shared gated SAE representation and introduces two technical contributions over that representation: governance-conditioned downstream sparse feature selection and decision-linked within-video evidence ranking.

The first technical contribution is governance-conditioned downstream sparse feature selection. SPACE learns a global relevance gate over the pretrained SAE feature space through the joint prediction, localization, policy-alignment, and sparsity objectives, and then converts the learned relevance structure into a discrete deployment vocabulary using Pareto-efficient selection and the one-standard-error rule. The learned gate determines which SAE features are retained, while the Pareto 1-SE rule determines the size of the deployment vocabulary. Our methodological contribution lies in deriving a compact, task-specific evidence vocabulary from the pretrained SAE feature space while preserving direct traceability between individual feature contributions and model predictions. Matched-size comparisons show that predictive performance declines when the selected vocabulary is replaced with activation-based or random feature subsets of the same size, indicating that SPACE’s predictive performance depends on the task relevance of the selected features rather than vocabulary size alone.

The second technical contribution is decision-linked within-video evidence ranking. In harmful short videos, decision-relevant content may appear only briefly, and video-level supervision does not ensure that the prediction is supported by the harmful segment. SPACE derives segment harmfulness from the decomposable category scores used for prediction and applies within-video ranking so that annotated harmful segments carry stronger evidence than benign segments from the same video. Because the ranking objective operates directly on the decomposable category scores, it localizes evidence without breaking its mathematical connection to prediction. The ablation results are consistent with this coupling contributing to both evidence localization and predictive performance.

Following the design-research pathway described by Abbasi et al. (2024), these technical mechanisms can be abstracted beyond the SPACE implementation by identifying the problem characteristics that made them necessary. AI design research gains broader value when domain adaptation reveals how characteristics of the problem require particular artifact capabilities. In our setting, three characteristics are central. Decision-relevant evidence can be locally concentrated within a larger information object. The available interpretable feature space can be too large and too general for direct organizational use. The resulting algorithmic decision must be interpreted against an external governance standard. Table 11 abstracts the SPACE design into three corresponding design principles (DP). Across the three principles, localization, task adaptation, and policy grounding should operate on the same prediction-linked evidence substrate and jointly shape the predictive artifact. The literature-grounded design requirements summarized in Table 1 map onto these principles as follows: DR2 (evidence localization) informs DP1, DR3 (policy grounding) informs DP2, and DR4 (evidence inspectability) informs DP3; DR1 (computational traceability) serves as a cross-cutting architectural constraint across all three principles, requiring the localized, selected, and policy-grounded evidence to directly compose the prediction.

**Table 11. Design Principles Abstracted from SPACE**

| Design principle / Related DR(s) | Problem condition | Design prescription | SPACE instantiation and supporting evidence |
|---|---|---|---|
| **DP1 (DR1 + DR2)**: Localize prediction-linked evidence at the level where the decision-relevant signal occurs. | A decision about a larger information object depends on evidence concentrated in only part of that object, and relevant and irrelevant content can coexist within it. | Design the artifact to learn evidence at the local unit of analysis and apply localization supervision directly to the same evidence scores that compose the prediction. Evaluate evidence localization separately from predictive correctness. | SPACE applies segment-level supervision and within-video ranking directly to decomposable category scores. Full SPACE achieves a Top-1 Evidence Hit Rate of 0.8800; removing within-video ranking lowers the hit rate to 0.8367. |
| **DP2 (DR1 + DR3):** Ground prediction-linked computational evidence in the external standard that governs the decision. | An algorithmic prediction informs an organizational action whose justification depends on an external policy, rule, guideline, or other decision standard. | Preserve the connection between the reported evidence and the prediction it composes, and align the shared evidence representation with an explicit external governance standard. Evaluate policy mapping separately from predictive performance. | SPACE aligns the pooled video evidence representation with category-level policy descriptions and achieves a Policy Mapping Accuracy of 0.8700. The present implementation supports category-level policy grounding, not feature-to-clause or clause-level correspondence. |
| **DP3 (DR1 + DR4):** Adapt general-purpose interpretable representations into a bounded, task-specific evidence vocabulary. | A pretrained model provides a large interpretable feature space, but only a subset of those features is relevant to the downstream task and the complete feature space is too large for practical inspection. | Learn governance-conditioned relevance over the interpretable feature space, select a discrete evidence vocabulary using explicit predictive-utility and parsimony criteria, and examine whether the retained features exhibit task-relevant semantic structure. | SPACE learns a global relevance gate over 131,072 SAE features and selects the deployment vocabulary through Pareto-efficient evaluation and the one-standard-error rule. Category-specific activation patterns and top-activating observations are used to examine the task relevance and semantic structure of the retained features. Replacing the selected vocabulary with a matched-size random vocabulary under the fixed-parameter substitution test in Table 8 lowers harmful-category Macro F1 from 0.8507 to 0.5091. |

These design principles specify the conditions under which the design contribution may transfer beyond the present setting. DP1 applies when evidence is locally concentrated within a larger object. DP3 applies when a general-purpose interpretable representation must be converted into a manageable, task-specific evidence space. DP2 applies when the artifact's output must be justified with reference to an external governance standard. Thus, the transferable contribution resides in this design logic, whereas the specific models, representations, and optimization methods may vary across settings.

For example, in cybersecurity incident detection, DP1 would localize the few anomalous log events that compose an incident-risk score; DP3 would convert a large, general-purpose interpretable feature space into a bounded, task-specific evidence vocabulary of attack indicators; and DP2 would ground the same prediction-linked evidence in organizational security policies or incident-response standards. Similarly, in automated credit underwriting, DP1 would localize decisive evidence within a broader application and transaction history; DP3 would retain a bounded, task-specific evidence vocabulary; and DP2 would connect that evidence to documented underwriting criteria and adverse-action rules.

These examples illustrate how the same design logic may be implemented differently across settings while preserving the functions specified by the three principles. SPACE provides one such implementation for toxic diet culture and may be adapted to other harmful-content moderation domains that exhibit the same conditions. Such adaptation requires domain-specific data, policy descriptions, and validation. More broadly, automated moderation becomes more governable when a system connects the relevant content, the computational basis of the prediction, and the policy basis of the judgment in an inspectable evidence record for stakeholder review.

### 5.2 Practical Implications

For platforms and moderation teams, SPACE can identify potentially harmful videos and support focused review of the evidence underlying each prediction. The evidence report identifies the video segment carrying the strongest evidence, the sparse features contributing to the prediction, and the policy category associated with the shared evidence representation. A moderator can use this information to move directly to the relevant portion of the video, inspect the computational basis of the prediction, and record the information used in the review. The platform-professional panel strongly supported these uses, particularly for helping moderators focus on key frames, reducing workload, and supporting the recording, auditing, and explanation of decisions. These findings indicate that platform professionals perceived the SPACE evidence report as useful for moderation and audit-related activities within platform governance processes.

The report also changes the information available to content creators after a moderation decision. A generic warning may communicate that content has been restricted without showing which part of the video produced the concern. SPACE provides a more specific account by identifying the relevant observation and linking the decision to a policy category. Content creators responded especially positively to the greater transparency of the report, its support for identifying harmful segments, and its usefulness for deciding how a video could be revised before posting. These responses indicate that content creators perceived localized, policy-grounded evidence as useful for understanding what triggered a moderation decision and for receiving actionable feedback about possible revisions. An important consideration in translating this benefit into practice is that greater transparency may create a governance tradeoff: highly detailed disclosure could enable strategic editing that avoids detection without addressing the underlying policy concern. To reduce this risk, operational deployment can differentiate the level of disclosure by stakeholder role. Platform moderation and audit functions can retain the full localized, feature-level evidence report, while creator-facing reports can emphasize the applicable policy category and provide coarser localization or revision guidance.

### 5.3 Limitations and Future Research

The empirical setting of this study limits the scope of its findings. Specifically, the evaluation focuses on TikTok videos related to toxic diet culture, and the categories of harmful content are defined according to the platform's policies. Although the proposed design principles may apply to other settings with similar governance characteristics, their applicability across platforms and content domains requires further examination. Platforms vary in their policy categories, enforcement thresholds, available content signals, and moderation procedures. Future research should therefore evaluate the proposed mechanisms under different community standards and across other forms of harmful content. Such research can help determine which aspects of the design are broadly applicable across governance environments and which are specific to the policy framework and content domain examined in this study.

The current implementation also has several limitations concerning the granularity of video evidence and policy grounding. First, SPACE's use of temporal annotations also requires additional annotation effort beyond the video-level category labels used to fine-tune the LMM benchmarks. Second, each video is represented by eight uniformly sampled visual observations. This design has two limitations: (1) because SPACE does not incorporate audio narration or a separate transcript, it may not recover harmful evidence conveyed solely through speech without corresponding visual cues (we did not observe clear audio-only cases during data annotation, although modality-specific evidence was not separately annotated); (2) uniform sampling provides coarser temporal coverage for longer videos and may miss harmful visual or on-screen textual evidence between sampled observations. Future research could incorporate additional video modalities and examine denser or adaptive temporal sampling strategies to determine whether they further improve evidence localization. Third, the current implementation provides category-level policy grounding by aligning a pooled video evidence representation with category-level policy descriptions; future research could extend this design to clause-level and action-level grounding. Similarly, the feature-category heatmap and top-activating examples provide descriptive evidence of task-related structure and semantic inspectability among the displayed features.

The stakeholder evaluation provides evidence of the artifact's perceived governance utility but does not assess its realized outcomes in an operational moderation environment. In addition, the small platform-professional panel limits the strength of conclusions that can be drawn for this stakeholder group. Future research could evaluate the artifact in field settings to examine how evidence reports affect review time, decision consistency, false-positive resolution, appeal handling, audit quality, and creator responses. Future studies could also investigate the artifact's performance as community guidelines evolve. This is important because the value of policy alignment depends on the evidence system's ability to remain reliable when the external standards it applies are revised.

### 5.4 Conclusion

Automated content moderation increasingly requires systems that can support the decision made after a prediction, not simply produce the prediction itself. SPACE shows that harmful-video detection can maintain predictive performance while identifying where decision-relevant content appears, tracing the prediction to a compact set of contributing internal features, and grounding that evidence in a platform policy category. The study therefore places evidence generation inside the design of the predictive artifact. As algorithmic predictions become more closely tied to platform governance, systems will increasingly need to expose inspectable evidence that can support and justify the moderation decisions those predictions inform.

## REFERENCES

Abbasi, A., Parsons, J., Pant, G., Sheng, O. R. L., & Sarker, S. (2024). Pathways for Design Research on Artificial Intelligence. *Information Systems Research*, *35*(2), 441–459. https://doi.org/10.1287/isre.2024.editorial.v35.n2

Alter, C. (2025, June 18). She Says Social-Media Algorithms Led to Her Eating Disorder. Now She's Suing TikTok and Instagram. *TIME*. https://time.com/7295323/social-media-case-instagram-tiktok/

Bai, S., Cai, Y., Chen, R., Chen, K., Chen, X., Cheng, Z., Deng, L., Ding, W., Gao, C., Ge, C., Ge, W., Guo, Z., Huang, Q., Huang, J., Huang, F., Hui, B., Jiang, S., Li, Z., Li, M., … Zhu, K. (2025). *Qwen3-VL Technical Report* (arXiv:2511.21631). arXiv. https://doi.org/10.48550/arXiv.2511.21631

Bertasius, G., Wang, H., & Torresani, L. (2021). Is Space-Time Attention All You Need for Video Understanding? *Proceedings of the 38th International Conference on Machine Learning*, 813–824. https://proceedings.mlr.press/v139/bertasius21a.html

Chancellor, S., Kalantidis, Y., Pater, J. A., De Choudhury, M., & Shamma, D. A. (2017). Multimodal Classification of Moderated Online Pro-Eating Disorder Content. *Proceedings of the 2017 CHI Conference on Human Factors in Computing Systems, CHI '17*, 3213–3226. https://doi.org/10.1145/3025453.3025985

Cherti, M., Beaumont, R., Wightman, R., Wortsman, M., Ilharco, G., Gordon, C., Schuhmann, C., Schmidt, L., & Jitsev, J. (2023). Reproducible Scaling Laws for Contrastive Language-Image Learning. *2023 IEEE/CVF Conference on Computer Vision and Pattern Recognition (CVPR)*, 2818–2829. https://doi.org/10.1109/CVPR52729.2023.00276

Cunningham, H., Ewart, A., Riggs, L., Huben, R., & Sharkey, L. (2023). *Sparse Autoencoders Find Highly Interpretable Features in Language Models* (arXiv:2309.08600). arXiv. https://doi.org/10.48550/arXiv.2309.08600

Dokme, A., & Vishwanath, S. (2026). *Interpreting Video Representations with Spatio-Temporal Sparse Autoencoders*. https://doi.org/10.1145/3767308.3836082

Gallifant, J., Chen, S., Sasse, K., Aerts, H., Hartvigsen, T., & Bitterman, D. (2025). Sparse Autoencoder Features for Classifications and Transferability. In C. Christodoulopoulos, T. Chakraborty, C. Rose, & V. Peng (Eds.), *Proceedings of the 2025 Conference on Empirical Methods in Natural Language Processing* (pp. 29939–29963). Association for Computational Linguistics. https://doi.org/10.18653/v1/2025.emnlp-main.1521

Google DeepMind. (2026). *Gemini 3.1 Pro*. Google DeepMind. https://deepmind.google/models/gemini/pro/

Gregor, S., & Hevner, A. R. (2013). Positioning and Presenting Design Science Research for Maximum Impact. *Management Information Systems Quarterly*, *37*(2), 337–355. https://doi.org/10.25300/MISQ/2013/37.2.01

He, H., Zhou, J., Shang, S., Hu, Y., Zhang, Y., & Zhou, K. (2026). *VideoOdyssey: A Benchmark for Ultra-Long-Context and Omni-Modal Video Understanding* (arXiv:2605.22907). arXiv. https://doi.org/10.48550/arXiv.2605.22907

He, Q., Hong, Y., & Raghu, T. S. (2025). Platform Governance with Algorithm-Based Content Moderation: An Empirical Study on Reddit. *Information Systems Research*, *36*(2), 1078–1095. https://doi.org/10.1287/isre.2021.0036

Hevner, A. R., March, S. T., Park, J., & Ram, S. (2004). Design science in information systems research. *Management Information Systems Quarterly*, *28*(1), 75–105. https://doi.org/10.2307/25148625

Holland, G., & Tiggemann, M. (2016). A systematic review of the impact of the use of social networking sites on body image and disordered eating outcomes. *Body Image*, *17*, 100–110. https://doi.org/10.1016/j.bodyim.2016.02.008

Kantamneni, S., Engels, J., Rajamanoharan, S., Tegmark, M., & Nanda, N. (2025). Are Sparse Autoencoders Useful? A Case Study in Sparse Probing. *Proceedings of the 42nd International Conference on Machine Learning*, 29018–29049. https://proceedings.mlr.press/v267/kantamneni25a.html

Kim, S.-S., Kim, S., & Kim, H.-W. (2025). Hate Speech Detection on Online News Platforms: A Deep-Learning Approach Based on Agenda-Setting Theory. *Journal of Management Information Systems*, *42*(3), 673–705. https://doi.org/10.1080/07421222.2025.2520173

Landis, J. R., & Koch, G. G. (1977). The Measurement of Observer Agreement for Categorical Data. *Biometrics*, *33*(1), 159–174. https://doi.org/10.2307/2529310

Lang, J., Hong, R., Zhong, M., Li, K., Zhong, T., Gao, Q., & Zhou, F. (2026). LEAF: Towards Lightweight Explainable Hateful Video Detection via Self-Grounding CoT Guided Stage-Wise Distillation. In M. Liakata, V. P. Moreira, J. Zhang, & D. Jurgens (Eds.), *Findings of the Association for Computational Linguistics: ACL 2026* (pp. 12420–12438). Association for Computational Linguistics. https://doi.org/10.18653/v1/2026.findings-acl.604

Li, K., Hong, R., Lang, J., Wu, J., Zhou, F., & Song, J. (2026). MATCH: Multi-Agentic Evidence Grounding for Explainable Hate Video Detection. *IEEE Transactions on Circuits and Systems for Video Technology*, *36*(7), 9646–9659. https://doi.org/10.1109/TCSVT.2026.3672052

Liu, M., Sun, Y., Sun, R., Dong, X., Shen, X., Wang, H., Xiong, H., & Song, Y. (2026). IPS: In-Prompt Process Supervision for Short Video Content Moderation. In Y. Li, G. Rehm, & M. Tu (Eds.), *Proceedings of the 64th Annual Meeting of the Association for Computational Linguistics (Volume 6: Industry Track)* (pp. 1277–1288). Association for Computational Linguistics. https://doi.org/10.18653/v1/2026.acl-industry.89

Lundberg, S. M., & Lee, S.-I. (2017). A Unified Approach to Interpreting Model Predictions. *Advances in Neural Information Processing Systems*, *30*. https://proceedings.neurips.cc/paper_files/paper/2017/hash/8a20a8621978632d76c43dfd28b67767-Abstract.html

Ni, B., Peng, H., Chen, M., Zhang, S., Meng, G., Fu, J., Xiang, S., & Ling, H. (2022). Expanding Language-Image Pretrained Models for General Video Recognition. *Computer Vision – ECCV 2022: 17th European Conference, Tel Aviv, Israel, October 23–27, 2022, Proceedings,*

*Part IV*, 1–18. https://doi.org/10.1007/978-3-031-19772-7_1
OpenAI. (2026, March 5). *Introducing GPT-5.4*. OpenAI. https://openai.com/index/introducing-gpt-5-4/
Padmanabhan, B., Fang, X., Sahoo, N., & Burton-Jones, A. (2022). Editor's Comments: Machine Learning in Information Systems Research. *Management Information Systems Quarterly*, *46*(1), iii–xix. https://doi.org/10.25300/MISQ/2022/461E1
Paul, K. (2021, October 16). 'It spreads like a disease': How pro-eating-disorder videos reach teens on TikTok. *The Guardian*. https://www.theguardian.com/technology/2021/oct/16/tiktok-eating-disorder-thinspo-teens
Rai, A. (2017). Editor's Comments: Diversity of Design Science Research. *Management Information Systems Quarterly*, *41*(1), iii–xviii.
Rajamanoharan, S., Conmy, A., Smith, L., Lieberum, T., Varma, V., Kramár, J., Shah, R., & Nanda, N. (2024). *Improving Dictionary Learning with Gated Sparse Autoencoders* (arXiv:2404.16014). arXiv. https://doi.org/10.48550/arXiv.2404.16014
Regulation (EU) 2022/2065. (2022, October 19). *Regulation (EU) 2022/2065 of the European Parliament and of the Council of 19 October 2022 on a Single Market For Digital Services and amending Directive 2000/31/EC (Digital Services Act) (Text with EEA relevance)*. http://data.europa.eu/eli/reg/2022/2065/oj
Ribeiro, M. T., Singh, S., & Guestrin, C. (2016). "Why Should I Trust You?": Explaining the Predictions of Any Classifier. *Proceedings of the 22nd ACM SIGKDD International Conference on Knowledge Discovery and Data Mining, KDD '16*, 1135–1144. https://doi.org/10.1145/2939672.2939778
Rudin, C. (2019). Stop explaining black box machine learning models for high stakes decisions and use interpretable models instead. *Nature Machine Intelligence*, *1*(5), 206–215. https://doi.org/10.1038/s42256-019-0048-x
Samory, M., Pamfile, D., To, A., & Phadke, S. (2026). Asking for It: Question-Answering for Predicting Rule Infractions in Online Content Moderation. *Proceedings of the International AAAI Conference on Web and Social Media*, *20*(1), 1993–2005. https://doi.org/10.1609/icwsm.v20i1.42734
Song, X., Muhamed, A., Zheng, Y., Kong, L., Tang, Z., Diab, M. T., Smith, V., & Zhang, K. (2026). Mechanistic Interpretability Should Prioritize Feature Consistency in Sparse Autoencoders. In M. Liakata, V. P. Moreira, J. Zhang, & D. Jurgens (Eds.), *Proceedings of the 64th Annual Meeting of the Association for Computational Linguistics (Volume 1: Long Papers)* (pp. 2172–2210). Association for Computational Linguistics. https://doi.org/10.18653/v1/2026.acl-long.99
Sultani, W., Chen, C., & Shah, M. (2018). *Real-World Anomaly Detection in Surveillance Videos*. 6479–6488. https://openaccess.thecvf.com/content_cvpr_2018/html/Sultani_Real-World_Anomaly_Detection_CVPR_2018_paper.html
Suzor, N. P., West, S. M., Quodling, A., & York, J. (2019). What Do We Mean When We Talk About Transparency? Toward Meaningful Transparency in Commercial Content Moderation. *International Journal of Communication*, *13*, 18–18.
TikTok. (2024, April 17). *TikTok's Mental and Behavioral Health policy*. https://www.tiktok.com/safety/en-GB/policies-and-engagement/mental-behavioral-health?cgversion=2025H2update
Tran, D., Wang, H., Torresani, L., Ray, J., LeCun, Y., & Paluri, M. (2018). A Closer Look at Spatiotemporal Convolutions for Action Recognition. *2018 IEEE/CVF Conference on Computer Vision and Pattern Recognition*, 6450–6459. https://doi.org/10.1109/CVPR.2018.00675
Wang, L., Huang, B., Zhao, Z., Tong, Z., He, Y., Wang, Y., Wang, Y., & Qiao, Y. (2023). VideoMAE V2: Scaling Video Masked Autoencoders with Dual Masking. *2023 IEEE/CVF Conference on Computer Vision and Pattern Recognition (CVPR)*, 14549–14560. https://doi.org/10.1109/CVPR52729.2023.01398
Wang, Y., Li, X., Yan, Z., He, Y., Yu, J., Zeng, X., Wang, C., Ma, C., Huang, H., Gao, J., Dou, M., Chen, K., Wang, W., Qiao, Y., Wang, Y., & Wang, L. (2025). *InternVideo2.5: Empowering Video MLLMs with Long and Rich Context Modeling* (arXiv:2501.12386). arXiv. https://doi.org/10.48550/arXiv.2501.12386
Wei, D., Chau, M., & Li, Z. (2025). Mitigating Bias in Hate Speech Detection with a Small Number of Expert Annotations: A Prompt-Based Learning Approach. *Management Information Systems Quarterly*, *49*(4), 1483–1512. https://doi.org/10.25300/MISQ/2025/18416
Xie, J., Chai, Y., Liang, R., Liu, Y., & Zeng, D. D. (2026). Short-Form Videos and Mental Health: A Knowledge-Guided Neural Topic Model. *Information Systems Research*, *37*(1), 356–377. https://doi.org/10.1287/isre.2024.1071
Yadav, A., & Singh, V. (2026). An Interpretable Agentic Framework for Multimodal Hate Video Analysis with Explicit Evidence Attribution. *Companion Proceedings of the ACM Web Conference 2026, WWW Companion '26*, 464–472. https://doi.org/10.1145/3774905.3796488
Yang, K., Zhang, Y., Du, M., Zhang, Y., Zheng, D., Zhao, K., Xiao, Y., Liang, H., & Xiao, K. (2026). UNIVID: Unified Vision-Language Model for Video Moderation. In Y. Li, G. Rehm, & M. Tu (Eds.), *Proceedings of the 64th Annual Meeting of the Association for Computational Linguistics (Volume 6: Industry Track)* (pp. 467–479). Association for Computational Linguistics. https://doi.org/10.18653/v1/2026.acl-industry.32
Zhang, B., Li, K., Cheng, Z., Hu, Z., Yuan, Y., Chen, G., Leng, S., Jiang, Y., Zhang, H., Li, X., Jin, P., Zhang, W., Wang, F., Bing, L., & Zhao, D. (2025, January 22). *VideoLLaMA 3: Frontier Multimodal Foundation Models for Image and Video Understanding*. arXiv.Org. https://arxiv.org/abs/2501.13106v4
Zhang, K., Shen, Y., Li, B., & Liu, Z. (2025). Large Multi-modal Models Can Interpret Features in Large Multi-modal Models. *Proceedings of the IEEE/CVF International Conference on Computer Vision*, 3650–3661. https://doi.org/10.1109/ICCV51701.2025.00348
Zhang, X., Wei, Z., Du, Q., & Zhang, Z. (2026). Social Media Moderation and Content Generation: Evidence From User Bans. *Management Information Systems Quarterly*, *50*(1), 211–242. https://doi.org/10.25300/MISQ/2025/18108